%% file: main.tex
\documentclass[lettersize,journal]{IEEEtran}
\usepackage{amsmath,amsfonts}
\usepackage{algorithmic}
\usepackage{algorithm}
\usepackage{array}
\usepackage[caption=false,font=normalsize,labelfont=sf,textfont=sf]{subfig}
\usepackage{textcomp}
\usepackage{stfloats}
\usepackage{url}
\usepackage{verbatim}
\usepackage{graphicx}
\usepackage{cite}

\usepackage{multirow}
\usepackage[figuresright]{rotating}
\usepackage{bigstrut}
\usepackage{amssymb}
\usepackage{booktabs} 
\usepackage[colorlinks, linkcolor=red, urlcolor=magenta, citecolor=blue]{hyperref}
\usepackage[numbers]{natbib}
\usepackage{makecell}
\usepackage{mathtools}
\usepackage{graphicx}
\usepackage{amssymb}
\usepackage{array}
\usepackage{epstopdf}
\usepackage{adjustbox}
\usepackage{xcolor}
\usepackage{diagbox}
\usepackage{amsfonts,amssymb} 
\usepackage{bbm}

\begin{document}

\title{Unsupervised Representation Learning for Time Series: A Review}

\author{Qianwen Meng,
        Hangwei Qian,
        Yong Liu,
        Yonghui Xu,
        Zhiqi Shen,
        and~Lizhen Cui,~\IEEEmembership{Senior Member,~IEEE}
\thanks{This work was supported in part by the National Key R\&D Program of China No. 2021YFF0900800; the Major Scientific and Technological Innovation Project of Shandong Province No. 2021CXGC010108; the Natural Science Foundation of Shandong Province No. ZR202111180007; the Fundamental Research Fund of Shandong University, and the Chinese Government Scholarship - High Level Postgraduate Program of China Scholarship Council.~\textit{(Corresponding authors: Hangwei Qian; Lizhen Cui.)}}  

\thanks{Q. Meng is with the School of Software, Shandong University, Jinan, China, and Joint SDU-NTU Centre for Artificial Intelligence Research, Shandong University, Jinan, China (Email: mqw\_sdu@mail.sdu.edu.cn).}
\thanks{H. Qian is with Centre for Frontier AI Research (CFAR), A*STAR, Singapore (Email: qian\_hangwei@cfar.a-star.edu.sg).}
\thanks{Y. Liu is with Joint NTU-UBC Research Centre of Excellence in Active Living for the Elderly (LILY), Nanyang Technological University, Singapore (Email: stephenliu@ntu.edu.sg).}
\thanks{Y. Xu is with the School of Software, Shandong University, Jinan, China and Joint SDU-NTU Centre for Artificial Intelligence Research, Shandong University, Jinan, China (Email: xuyonghui@sdu.edu.cn).}
\thanks{Z. Shen is with the School of Computer Science and Engineering, Nanyang Technological University, Singapore (Email: zqshen@ntu.edu.sg).}
\thanks{L. Cui is with the School of Software, Shandong University, Jinan, China and Joint SDU-NTU Centre for Artificial Intelligence Research, Shandong University, Jinan, China (Email: clz@sdu.edu.cn).}

}
\markboth{Journal of \LaTeX\ Class Files,~Vol.~14, No.~8, August~2021}%
{Shell \MakeLowercase{\textit{et al.}}: A Sample Article Using IEEEtran.cls for IEEE Journals}


\maketitle

\begin{abstract}
Unsupervised representation learning approaches aim to learn discriminative feature representations from unlabeled data, without the requirement of annotating every sample. Enabling unsupervised representation learning is extremely crucial for time series data, due to its unique annotation bottleneck caused by its complex characteristics and lack of visual cues compared with other data modalities. In recent years, unsupervised representation learning techniques have advanced rapidly in various domains. However, there is a lack of systematic analysis of unsupervised representation learning approaches for time series. To fill the gap, we conduct a comprehensive literature review of existing rapidly evolving unsupervised representation learning approaches for time series. Moreover, we also develop a unified and standardized library, named ULTS (\emph{i.e.,} Unsupervised Learning for Time Series), to facilitate fast implementations and unified evaluations on various models. With ULTS, we empirically evaluate state-of-the-art approaches, especially the rapidly evolving contrastive learning methods, on 9 diverse real-world datasets. We further discuss practical considerations as well as open research challenges on unsupervised representation learning for time series to facilitate future research in this field. 
\end{abstract}


\section{Introduction}
\IEEEPARstart{T}{ime} series data can record the changing trends of variables over time and are ubiquitous across various fields such as Internet of Things (IoT)~\cite{iot1,iot2,sde_iot}, industry 4.0~\cite{industry1,industry2}, financial services~\cite{finance1}, and healthcare services~\cite{medical1,medical2}. In order to extract and infer useful information from complex raw time series data, time series representation learning has emerged as an effective learning paradigm. On the one hand, the learned latent representations can capture potentially valuable information within time series and reveal the underlying mechanisms of the corresponding systems or phenomena. On the other hand, high-quality representations are essential for numerous downstream tasks, including time series classification~\cite{mts_clf,evaluation_clf_17,evaluation_clf_21}, forecasting~\cite{survey_forecasting_dl,survey_forecasting_semiml,evaluation_forecasting,sde_prediction}, and clustering~\cite{ts_clustering,isea}. Pervasive raw time series data are collected through a wide spectrum of sensors or wearable devices with varying frequencies~\cite{sde_sensor}. Annotating corresponding labels for such data, however, is inevitably inefficient and could be erroneous due to the lack of obvious visual patterns in the raw data. This would inhibit the training of conventional supervised learning models since full label annotations are required beforehand. Weakly-supervised learning methods can alleviate the annotation challenge to some extent, but they still necessitate partial or coarse labels for model training~\cite{DBLP:journals/ai/QianPM21}. In this paper, we turn to the techniques that aim to learn high-quality representations from raw time series data in an unsupervised learning manner.

\begin{table*}[htbp]
\renewcommand{\arraystretch}{0.9}
\setlength{\tabcolsep}{5pt}
\centering
\small
      \begin{tabular}{c|ccc|cccc|cc|p{10cm}}
      \toprule[1.2pt]
      \multicolumn{1}{c|}{\multirow{6}[4]{*}{ Surveys}} & \multicolumn{3}{c|}{Applications} & \multicolumn{4}{c|}{Core Techniques} &\multicolumn{2}{c|}{Types}& \multicolumn{1}{c}{\multirow{6}[4]{*}{ Main Contributions}} \\
\cline{2-10}   & \multicolumn{1}{c}{\begin{sideways}\textbf{Classification}\end{sideways}} & \multicolumn{1}{c}{\begin{sideways}Forecasting\end{sideways}} & \multicolumn{1}{c|}{\begin{sideways}Others\end{sideways}} & \multicolumn{1}{c}{\begin{sideways}Supervised\end{sideways}} & \multicolumn{1}{c}{\begin{sideways}\textbf{Unsupervised}\end{sideways}} & \multicolumn{1}{c}{\begin{sideways}\textbf{Deep}\end{sideways}} & \multicolumn{1}{c|}{\begin{sideways}Non-deep\end{sideways}} & \multicolumn{1}{c}{\begin{sideways}Univariate\end{sideways}} & \multicolumn{1}{c|}{\begin{sideways}\textbf{Multivariate}\end{sideways}} &  \\
      \midrule
      ~\cite{evaluation_clf_17}    & \checkmark  &   &   & \checkmark  &   &   & \checkmark  & \checkmark  &   &  They classify algorithms according to the type of discriminatory features the technique is attempting to find. The algorithms are grouped into those that use the whole series, intervals, shapelets or repeating pattern counts.\\
      \midrule
    ~\cite{evaluation_clf_21}  & \checkmark  &   &   & \checkmark  &   & \checkmark  & \checkmark  &  & \checkmark  &  They review recently proposed bespoke multivariate time series classification algorithms based on deep learning, shapelets and bag-of-words approaches. \\
    \midrule
   ~\cite{review_clf_dl}  &  \checkmark &   &   &  \checkmark &   &  \checkmark &   &  \checkmark &  \checkmark &  They group deep learning approaches for time series classification into generative models including auto-encoders and echo state networks, and discriminative models including feature engineering and end-to-end methods.\\
     \midrule
    ~\cite{review_clf_xai}    &  \checkmark &   &   &  \checkmark &   &  \checkmark &   & \checkmark  & \checkmark  &  They examine the current state of eXplainable AI (XAI) methods with a focus on approaches for opening up deep learning black boxes including MLPs, RNNs, CNNs and ResNets for the task of time series classification.
    \\
     \midrule
  ~\cite{review_clf_early}  & \checkmark &   &   &  \checkmark &   &  \checkmark & \checkmark  & \checkmark & \checkmark   &  They categorize early classification approaches for incomplete time series into 4 exclusive categories based on their proposed solution strategies, including prefix-based, shapelet-based, model-based and miscellaneous ones. \\
      \midrule
  ~\cite{review_clf_distance} &  \checkmark &   &   & \checkmark  &   &   & \checkmark  & \checkmark & \checkmark  &  They propose a taxonomy of distance-based classifiers, including KNN-based, distance features-based and distance kernels-based ones.\\
  \midrule
      ~\cite{evaluation_forecasting} &   & \checkmark  &   & \checkmark  &   &  \checkmark &   & \checkmark  &   &  They review 3 categories of deep learning architectures for time series forecasting, including MLPs, RNNs and CNNs.  \\
  \midrule
   ~\cite{survey_forecasting_dl} &   & \checkmark  &   &  \checkmark &   &  \checkmark &   & \checkmark & \checkmark  &  They classify deep learning methods for time series forecasting by application domains (energy and fuels, image and video, finance, industry, health, etc.) and network architectures (ENN, LSTM, GRU, BRNN, DFFNN, CNN, TCN). \\ 
   \midrule
   ~\cite{survey_ts_compression} &   &   & \checkmark  &  \checkmark  &   &  & \checkmark  &  \checkmark &  \checkmark &  They divide compression algorithms into dictionary-based methods, functional approximation, auto-encoders, and sequential algorithms.\\
   \midrule
   ~\cite{review_temporal}&   &    &  \checkmark &   & \checkmark   &  \checkmark   &   &   &  \checkmark  &  They review approaches specifically proposed for multi-modal and temporal data, then categorize discriminative self-supervised representation learning models into pretext, contrastive, clustering and regularisation-based models.\\
    \midrule
   ~\cite{survey_ts_mining} &   &   & \checkmark   & \checkmark   &   &   &  \checkmark  &   &   &  They divide main research orientations into 3 subfields: dimensionality reduction (time series representation), similarity measures and data mining tasks.\\
   \midrule
   ~\cite{langkvist2014review} &   &   &  \checkmark &   &  \checkmark & \checkmark  &   &  \checkmark &  \checkmark & They review a variety of feature learning algorithms that have been developed to explicitly capture temporal relationships (such as RBM, AE, RNN, TDNN).\\ 
    \hline
    Ours &  \checkmark &  \checkmark &  \checkmark  &   &  \checkmark & \checkmark  &   & \checkmark &  \checkmark &  Our comprehensive review encompasses a broad spectrum of unsupervised representation learning methods, including both established and novel approaches, with a particular emphasis on the latest advancements in contrastive learning techniques. Furthermore, we develop a unified and standardized library called ULTS, which serves as a valuable resource for convenient empirical evaluations and comparisons of these methods. \\
      \bottomrule[1.2pt]
      \end{tabular}
    \caption{Comparisons of existing related surveys on time series representation learning.}
    \vspace{-15pt}
    \label{tab:surveys}
  \end{table*}

In general, unsupervised representation learning aims to extract latent representations from complex raw time series data without human supervision~\cite{disentangledts}. Previous studies have shown remarkable performance with joint clustering and feature learning approaches~\cite{odc}. However, clustering primarily focuses on identifying specific data patterns and can be unreliable due to the poor generality of predefined priors~\cite{sde_clustering}. The advent of deep learning techniques has led to the widespread adoption of auto-encoders and seq-to-seq models for representation learning~\cite{survey_ts_seq2seq}. These approaches employ appropriate learning objectives, such as self-reconstruction and context prediction, to learn meaningful representations~\cite{ts_clustering,timenet}. Nevertheless, accurately reconstructing the complex time series data remains challenging, especially with the high-frequency physiological signals~\cite{tnc,sslecg}. Recently, self-supervised learning is emerging as a new paradigm, which induces supervision by designing pretext tasks instead of relying on pre-defined prior knowledge~\cite{review_unified_graph, survey_ptcl,survey_cl_ssl}. Diverging from fully unsupervised approaches, self-supervised learning methods leverage pretext tasks to autonomously generate labels by utilizing intrinsic information derived from the unlabeled data~\cite{evaluation_har_cl}. The learning efficiency can be improved through discriminative pre-training based on self-generated supervised signals that are freely obtained from the raw data~\cite{review_temporal,ssl1,ssl2,ssl3}.

More recently, self-supervised learning has made significant leaps fueled by advancements in contrastive learning, which employs the instance discrimination pretext task to bring similar pairs closer while pushing dissimilar pairs apart in the feature space~\cite{cl_false_cancel}. Contrastive learning has achieved remarkable advantages in representation learning for various types of data, including image~\cite{simclr,byol,moco,respnet}, video~\cite{tclr,video2,video3} and time series~\cite{tloss,tstcc,ts2vec,timesnet}. Alignment serves as a key motivation in representation learning methods~\cite{DBLP:conf/icml/0001I20}. Specifically, in the context of contrastive learning, alignment involves the mapping of 2 samples that form a positive pair to nearby features, resulting in a high degree of invariance to unneeded noise factors. Contrastive learning primarily focuses on 3 levels, namely instance-level contrast~\cite{simclr_2,cpc,simsiam}, prototype-level contrast~\cite{pcl,swav,mhccl}, and temporal-level contrast~\cite{tnc,tstcc,tcl}. The instance-level contrast treats samples independently and distinguishes them by pulling positive pairs together and pushing negative pairs apart~\cite{DBLP:conf/iclr/0001Z0YL22}. Prototype-level contrast, instead, goes beyond the independence assumption and exploits latent cluster information present within samples. Researchers have demonstrated that the learned representations are expected to retain higher-level semantic information by taking advantage of additional prior information brought by clustering~\cite{pcl,swav,idfd,ccl}. Furthermore, considering the temporal dependency inherent in time series data, researchers have explored the feasibility of distinguishing contextual information at a fine-grained temporal level~\cite{tloss,tstcc,ts2vec}.

As shown in Table~\ref{tab:surveys}, existing empirical surveys primarily focus on end-to-end models associated with specific downstream tasks such as time series classification~\cite{evaluation_clf_17,evaluation_clf_21,survey_ts_seq2seq,survey_ts_taxonomy}, forecasting~\cite{evaluation_forecasting,survey_forecasting_dl}, or anomaly detection~\cite{iot1,iot2}. However, they tend to overlook the crucial representation learning process that precedes these downstream tasks. The models reviewed in current surveys primarily prioritize the optimization of the entire end-to-end model, neglecting the in-depth exploration of learning meaningful representations. Consequently, there is a lack of systematic analysis of unsupervised representation learning methods that prioritize the acquisition of effective representations, whose efficacy on downstream tasks can be verified by appending simple classifiers or predictors on top of frozen representations. In addition, existing related surveys fail to cover the rapidly evolving self-supervised approaches, especially contrastive learning techniques. It can be observed that approximately half of existing surveys primarily focus on models that employ non-deep fundamental methods as their core techniques. Among the surveys that do incorporate deep learning models, the emphasis is mainly on supervised approaches. In addition, most of them exclusively cover traditional methods within a completely unsupervised setting, such as clustering~\cite{survey_ts_clustering,ts_clustering} or reconstruction-based methods~\cite{review_clf_dl,survey_ts_compression}, thereby overlooking numerous contemporary and highly effective techniques like self-supervised learning that have proven to yield superior results in recent years. To fill this gap, we carry out this extensive review of representation learning methods for time series from an unsupervised deep learning perspective.

Our comprehensive literature review encompasses \textit{\romannumeral1)} the deep clustering methods, \textit{\romannumeral2)} the reconstruction-based methods, and \textit{\romannumeral3)} the self-supervised learning methods particularly contrastive learning methods. The purpose of this review is to provide readers with a clear and thorough overview of cutting-edge research in the field of unsupervised time series representation learning. In addition, we observe that it is extremely difficult to conduct fair comparisons to evaluate existing works due to the rapid evolution and various taxonomies of unsupervised learning methods. These evaluations often differ in components such as augmentation methods, backbones, and even datasets, making it difficult to draw meaningful comparisons. Consequently, there is a lack of comprehensive benchmark evaluations that can offer researchers a holistic understanding of the strengths and weaknesses of existing approaches from various perspectives. We notice that the previous work~\cite{review_clf_dl} conducted an experimental evaluation of pure discriminative time series classification models, but excluded representation learning methods. In this work, we aim to bridge this gap by developing a unified and standardized library ULTS\footnote{The library \textbf{ULTS} is available at \url{https://github.com/mqwfrog/ULTS}.} to enable quick and convenient evaluations of unsupervised representation learning approaches for time series. Furthermore, we conduct empirical evaluations over state-of-the-art approaches on 9 diverse real-world datasets within this unified evaluation testbed. Our evaluation serves as a valuable reference for understanding the strengths and applicable scenarios of different models, helping researchers design and evaluate customized models in the future. Overall, the main contributions of this work are summarized as follows:
\begin{itemize}

\item We conduct a comprehensive literature review of unsupervised representation learning approaches for time series, and propose an up-to-date taxonomy that categorizes different approaches.

\item We develop a standardized library ULTS, to conveniently evaluate various models in a unified testbed, and conduct empirical evaluations of state-of-the-art methods, particularly contrastive ones, on 9 diverse real-world datasets.

\item We discuss practical considerations as well as open research challenges on unsupervised learning for time series and provide new insights to facilitate future research in this field. 

\end{itemize}

The rest of this paper is organized as follows: In Section~\ref{s2}, we introduce the definitions and terminology that are used throughout this paper, and propose a taxonomy that classifies existing unsupervised representation learning methods for time series. Additionally, we introduce a unified and standardized library ULTS that integrates 17 representative models encompassing various categories. Then, we briefly review related works including deep clustering, reconstruction-based and self-supervised learning methods in Section~\ref{s3}, Section~\ref{s4}, and Section~\ref{s5} respectively. Finally, we conclude this paper and discuss promising future research directions in Section~\ref{s6}.

\section{Overview}
\label{s2}
\subsection{Definitions and Terminology}
Time series representation learning holds utmost significance in time series modeling, as it strives to extract meaningful and informative latent representations within the reservoir space from the intricate raw complex time series data. The primary obstacle in time series representation learning lies in the lack of well-annotated labels, resulting in insufficient supervisory signals for training supervised models. Consequently, unsupervised learning approaches are devised to tackle this challenge by creating various pretext tasks that generate self-generated labels without relying on human supervision. We consider a set of $N$ time series $\mathbb{X}=\{\mathbf{x}_i\}_{i=1}^N$, where $\mathbf{x}_i \in \mathbb{R} ^ {T \times V }$ denotes a sequence of time series data collected from $V$ variables within $T$ timestamps. When $V = 1$, it becomes the univariate time series. In general, we can formulate the problem as follows: given an original time series $\mathbf{x}_i$, unsupervised representation learning aims to learn an encoder $f_{e}(\cdot)$ that can map $\mathbf{x}_i$ into a $D$-dimensional representation vector $\mathbf{z}_i$ in the latent space. A crucial measure to assess the quality and effectiveness of the acquired representations is their ability to serve subsequent downstream tasks. Consequently, once the latent representations are obtained, they are frozen without further modification, and then can be utilized as input for downstream tasks such as classification, forecasting, clustering, and anomaly detection to evaluate their performance. The unified mathematical notations used throughout this paper are summarized in Table~\ref{tab:notations}.

\begin{table}[htbp]
    \centering
    \small
     \resizebox{\columnwidth}{1.8cm}{
    \begin{tabular}{l|l}
    \toprule[1.2pt]
  \textbf{Notations}  &  \textbf{Descriptions} \\
    \midrule
  $\mathbb{X}$ & The set of original time series data\\
  $\mathbb{Z}$ & The set of learned representations in the latent space \\
  $\mathbb{C}$ & The set of cluster centroids (prototypes) \\
  $N$ & The number of samples\\
  $V$ & The number of variables\\
  $T$ & The number of timestamps (the length of time series)\\
  $f_{e}(\cdot)$ & The encoder that maps original data $\mathbb{X}$ to representations $\mathbb{Z}$\\
  $f_{d}(\cdot)$ & The decoder that uses representations $\mathbb{Z}$ to reconstruct original data $\mathbb{X}$\\
  $h(\cdot)$ & The nonlinear projection head to remap the representations \\
  $f_{m}(\cdot)$ & The momentum encoder \\
    \bottomrule[1.2pt]
    \end{tabular}
    }
    \caption{List of mathematical notations used throughout this paper.}
    \label{tab:notations}
\end{table}

\begin{figure*}[t]
  \centering
  \includegraphics[trim = 0mm 0mm 0mm 0mm, clip, width=\linewidth,height=9.3cm]{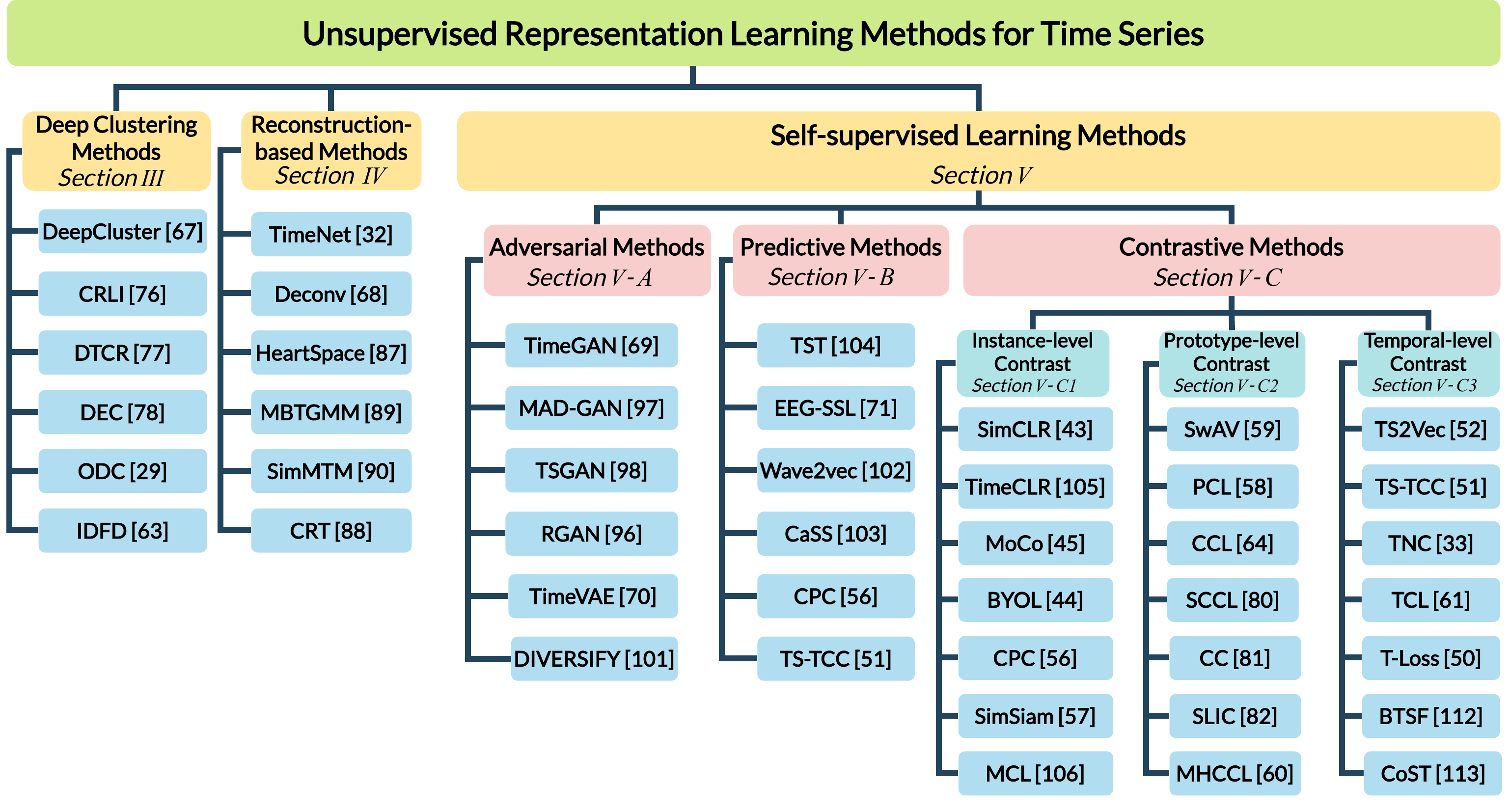}
  \caption{An up-to-date taxonomy of unsupervised representation learning methods for time series, including a) Deep Clustering Methods b) Reconstruction-based Methods and c) Self-supervised Learning Methods. The self-supervised learning methods can be further divided into adversarial methods, predictive methods and contrastive methods, depending on different types of pretext tasks employed for acquiring self-supervised signals. }
  \label{fig:taxonomy_tree}
\end{figure*}

\subsection{A Taxonomy of Unsupervised Representation Learning Methods for Time Series}

\begin{figure}[h]
  \centering 
  \includegraphics[trim = 0mm 0mm 0mm 0mm, clip, width=\columnwidth,height=5.7cm]{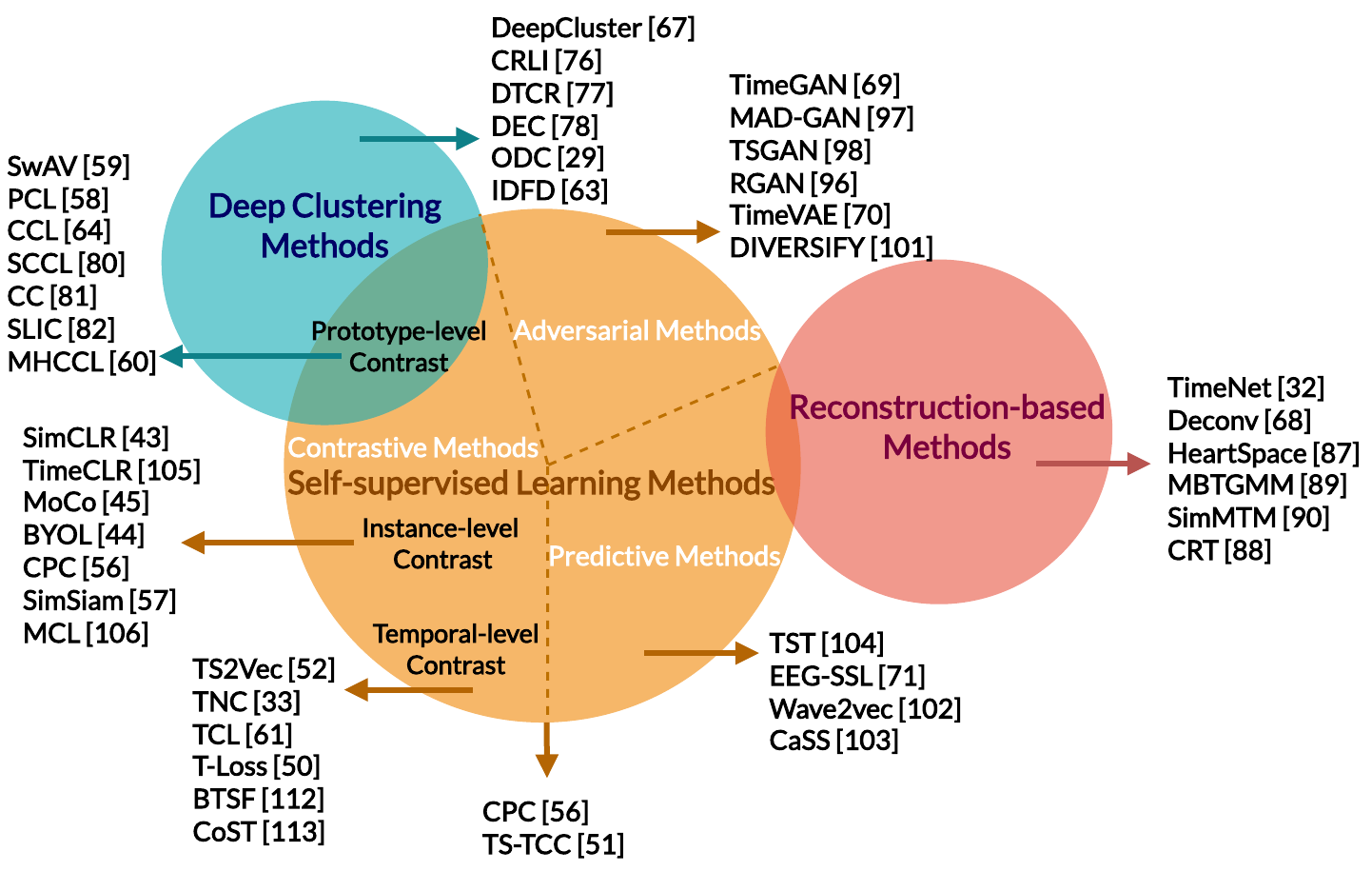}
  \caption{The organization of the research in unsupervised learning techniques for time series, based on 3 main categories outlined in our proposed taxonomy.}
  \label{fig:taxonomy_diagram}
\end{figure}

As presented in Figure~\ref{fig:taxonomy_tree}, we classify the existing unsupervised representation learning methods for time series into 3 categories: deep clustering, reconstruction-based, and self-supervised learning methods. Deep clustering methods typically combine traditional clustering techniques with deep learning neural networks, utilizing clustering results as labels for representation learning while updating clustering results based on learned representations. Reconstruction-based methods usually employ auto-encoders or seq-to-seq models to reconstruct either partial or the entire raw time series through joint training with decoders. More recently, self-supervised learning has gained popularity for extracting meaningful representations from unlabeled data as it avoids the need for manual annotation by designing pretext tasks. We further categorize self-supervised learning methods into adversarial, predictive, and contrastive ones, depending on different types of pretext tasks employed for acquiring self-supervised signals. Adversarial methods discriminate between fake and real data by using the discriminator, where the fake data similar to training data are produced by the generator. Predictive methods involve predicting future inputs from past inputs, predicting the original view from some other corrupted views, or predicting masked parts from unmasked parts. Contrastive methods perform distance discrimination using self-supervised signals generated from data augmentations. Specifically, contrastive methods can be subdivided into fine-grained subcategories according to different levels of contrast, including instance-level, prototype-level, and temporal-level approaches. Figure~\ref{fig:taxonomy_diagram} shows the organization of the research in unsupervised learning techniques for time series, based on the 3 main categories outlined in our proposed up-to-date taxonomy.

\subsection{The ULTS Library}
We have developed ULTS, a unified and standardized library under the PyTorch framework, to facilitate comparisons of existing models for unsupervised time series representation learning. ULTS integrates 17 state-of-the-art models, providing a unified testbed environment for evaluation. This library covers 2 deep clustering models, 2 reconstruction-based models, and 13 self-supervised learning models consisting of 2 adversarial models, 2 predictive models, and 9 contrastive models. Table~\ref{tab:library} lists all the models implemented in our ULTS library. ULTS also offers various flexible options for data augmentation transformations in both time and frequency domains. By default, it applies a combination of the weak augmentation (\emph{i.e.,} jitter-and-scale) and the strong augmentation (\emph{i.e.,} permutation-and-jitter), which aligns with the strategy used in TS-TCC~\cite{tstcc}. For more implementation details, please refer to the ULTS website. The comprehensive empirical experimental evaluation and analysis conducted by ULTS are provided in the Supplementary Materials.

\begin{table}[htb]
    \centering
    \setlength{\tabcolsep}{2.7pt}
    \renewcommand{\arraystretch}{1}
    \begin{tabular}{p{2.4cm}|c|c|l}
    \toprule[1.2pt]
  1st Category  &  2nd Category & 3rd Category &  Model \\
    \midrule
     \multirow{2}{3cm}{Deep Clustering} & - & - & DeepCluster~\cite{deepcluster} \\
     && & IDFD~\cite{idfd} \\
     \midrule
     \multirow{2}{*}{Reconstruction-based} & - & - & TimeNet~\cite{timenet} \\
     && & Deconv~\cite{deconv} \\
     \midrule
     \multirow{12}{3cm}{Self-supervised}  &  \multirow{2}{*}{Adversarial}& - & TimeGAN~\cite{timegan} \\
     &&  & TimeVAE~\cite{timevae} \\
     \cline{2-4}
     & \multirow{2}{*}{Predictive}  & -& EEG-SSL ~\cite{eegssl} \\
     && & TST~\cite{tst} \\
     \cline{2-4}
     & \multirow{9}{*}{Contrastive} & \multirow{3}{2.2cm}{Instance-Level}  &SimCLR~\cite{simclr} \\
      &&   & BYOL~\cite{byol} \\
      && & CPC~\cite{cpc} \\
      \cline{3-4}
    &  & \multirow{3}{2.2cm}{Prototype-Level}  &SwAV~\cite{swav} \\
     &&  & PCL~\cite{pcl} \\
      && & MHCCL~\cite{mhccl} \\
      \cline{3-4}
    &  &  \multirow{3}{2.2cm}{Temporal-Level}  &TS2Vec~\cite{ts2vec} \\
     & &  & TS-TCC~\cite{tstcc} \\
     & & & T-Loss~\cite{tloss} \\
    \bottomrule[1.2pt]
    \end{tabular}
    \caption{Models implemented in ULTS library.}
    \label{tab:library}
\end{table}

\section{Deep Clustering Methods}
\label{s3}
Joint clustering and feature learning approaches have shown remarkable performance in unsupervised representation learning~\cite{odc}. The primary objective is to discover underlying similar patterns and structures within data without relying on explicit labels or annotations. Traditional clustering algorithms such as flat clustering~\cite{DBLP:conf/cvpr/SarfrazSS19,ts_clustering}, hierarchical clustering~\cite{DBLP:conf/aaai/ShinSM20} and spectral clustering~\cite{DBLP:journals/ijcv/ZhangFWLCH20} essentially employ handcrafted feature extractors to group samples with similar characteristics together. However, these methods often exhibit limited performance when applied to time series data due to the complex and high-dimensional structures inherited therein. Simple distance-based metrics in traditional clustering struggle to effectively represent these intricate structures. Additionally, manually extracted features usually lack quality guarantees and the process is not automated. Deep neural networks have the ability to learn complex and hierarchical representations of data, making them well-suited for capturing the intricate nature of time series data.  In this case, the integration of traditional clustering techniques with deep learning, known as deep clustering, has garnered significant attention from researchers. Deep clustering methods break down the barriers between clustering and representation learning, allowing for the iterative optimization of the clustering-oriented objective to learn mappings from the input space to a new latent space~\cite{DBLP:journals/corr/abs-2206-07579, odc}. The iterative optimization approaches such as DeepCluster~\cite{deepcluster} and ODC~\cite{odc} enable the neural network and the clustering algorithm to mutually reinforce each other through interactive learning, leading to improved performance and more meaningful representations. CRLI~\cite{crli}, DTCR~\cite{dtcr} and DEC~\cite{dec} employ auto-encoder architectures as backbones and jointly optimize the reconstruction loss and clustering loss. In addition to the widely used K-Means clustering, spectral clustering is also explored by IDFD~\cite{idfd} to learn clustering-friendly representations. It is worth noting that clustering assignments can serve as self-supervised signals, thereby facilitating the self-learning process. Particularly, state-of-the-art contrastive learning highlights the importance of the alignment property that encourages the samples with similar features or semantic categories to be close in the low-dimensional space, which is also essential for deep clustering~\cite{DBLP:conf/iclr/0001YZJ23}. In this case, recent studies that leverage clustering-based contrastive learning to acquire highly effective representations with higher-level semantic information have emerged~\cite{swav,pcl,ccl,sccl,cc,slic, mhccl}, which will be discussed in Section~\ref{prototype-level cl}. More details of deep clustering methods for unsupervised representation learning are listed as follows.

\subsubsection{DeepCluster~\cite{deepcluster}} 
DeepCluster iteratively groups samples with the standard K-Means clustering algorithm, and uses the clustering assignments as supervised signals to update the weights of the network. Initially, the model extracts representations through the convolutional network. Subsequently, similar representations are grouped into the same class through clustering, and the resulting clustering labels are then utilized to train the classifier. However, such an iterative approach is susceptible to the accumulation of errors during the alternating stages of representation learning and clustering. Consequently, sub-optimal clustering performance may arise as a result.

\subsubsection{CRLI~\cite{crli}}
Clustering Representation Learning on Incomplete time series data (CRLI) jointly optimizes the imputation and clustering process to obtain more discriminant values for clustering, so that the learned representations possess good clustering property. In addition to reconstructing the original input time series, CRLI also encourages the learned representations to form cluster structures by integrating the standard soft K-means objective into the encoder-decoder network.

\subsubsection{DTCR~\cite{dtcr}}
Deep Temporal Clustering Representation (DTCR) integrates the temporal reconstruction and K-Means objective into the seq-to-seq model to generate cluster-specific temporal representations. To further enhance the capability of the encoder, DTCR proposes an auxiliary classification task and a fake-sample generation strategy involving the random shuffling of certain time steps. The model employs bidirectional dilated recurrent neural networks as the encoder, enabling the learned representations to effectively capture the temporal dynamics and multi-scale characteristics present in the original time series data.

\subsubsection{DEC~\cite{dec}}
Deep Embedded Clustering (DEC) achieves joint optimization by iteratively optimizing a KL divergence-based clustering objective with a centroid-based probability distribution. The underlying assumption of DEC is that the initial classifier’s highly confident predictions are mostly accurate. In contrast to DTCR~\cite{dtcr} that directly optimizes the K-Means process, DEC enhances the clustering performance following the initialization in the new feature space with a stacked auto-encoder(SAE) and standard K-Means.

\subsubsection{ODC~\cite{odc}}
Online Deep Clustering (ODC) devises a joint clustering and feature learning paradigm with high stability by performing clustering and network update simultaneously rather than performing them alternatively. To achieve this, ODC introduces novel techniques including loss re-weighting according to the number of samples in each class, and processing small clusters in advance to avoid ODC from getting stuck into trivial solutions. Loss re-weighting helps to prevent the formation of huge clusters and eliminates extremely small clusters in advance. Specifically, the samples and centroids' memories are initialized via K-Means clustering, and then the uninterrupted ODC is performed iteratively.

\subsubsection{IDFD~\cite{idfd}} 
Inspired by the properties of classical spectral clustering, IDFD applies deep representation learning via instance discrimination to clustering. In addition, IDFD introduces a softmax-formulated feature decorrelation constraint for learning presentations in the latent space to achieve stable improvements in clustering performance. The instance discrimination module is used to capture the similarities among data, while the feature decorrelation module is used to remove the redundant correlations among features.

\section{Reconstruction-based Methods}
\label{s4}
Compared to deep clustering methods that jointly learn representations and clustering assignments, reconstruction-based methods instead prioritize minimizing the discrepancy between the reconstructed output and the raw input, effectively training the network to disregard insignificant data that may contain noise~\cite{hinton2006reducing, yu2021analysis, s2sfa}. Reconstruction-based methods generally adopt utilize an encoder-decoder architecture to reconstruct the original input data from a modified or incomplete version of the input data. They learn meaningful representations by emphasizing the salient features and filtering out irrelevant or noisy information in the data~\cite{pointcloud}. Approaches such as TimeNet~\cite{timenet} and Deconv~\cite{deconv} reconstruct the whole time series in a reverse order to cater to the characteristics of time series data. Particularly, HeartSpace~\cite{heartspace} segments the heart rate data into day-long time series because human behavior presents day-long regularities. In addition, CRT~\cite{crt} and MBTGMM~\cite{mbtgmm} are able to capture long-term dependencies by utilizing transformer encoders compared to those that use convolutional neural networks~\cite{deconv, heartspace}. Unlike methods such as SimMTM~\cite{SimMTM} that mask certain segments of time series data for construction, CRT~\cite{crt} drops segments instead, preserving the original pattern and minimizing noise introduced during representation learning. The effectiveness of learned representations is evaluated by how accurately the decoder can regenerate the input data. Such evaluation ensures that the model captures and encodes crucial information from the input while maintaining the capacity to reconstruct it faithfully. Consequently, reconstruction can also be utilized as a pretext task into self-supervised learning since the generalization performance of pre-trained representations in downstream tasks can be further enhanced by reconstructing richer contextual information~\cite{DBLP:conf/iccv/ZhouLYHY21}. Below are additional details of reconstruction-based methods used for unsupervised representation learning.

\subsubsection{TimeNet~\cite{timenet}}
TimeNet tackles the challenges associated with time series data, including the varying lengths and the data scarcity. To reconstruct the time series, TimeNet employs a sequence auto-encoder (SAE) network based on the sequence-to-sequence model to transform the original time series of varying lengths into fixed-dimensional representations. The encoder RNN serves as a pre-trained model while the decoder RNN operates in reverse order, aligning with the inherent temporal nature of time series data. They are trained jointly to minimize the reconstruction error on time series.

\subsubsection{Deconv~\cite{deconv}}
Deconv utilizes deconvolutional networks that employ inverse operations of convolution and pooling to reconstruct time series data. This reverse procedure of convolution enables the reconstruction of hidden representations within the network. The output of traditional unpooling layers is usually an enlarged but sparse activation map, which may lead to a loss of expressiveness in complex features required for accurate reconstruction. To address this, Deconv combines unpooling and deconvolution techniques to effectively capture the cross-channel correlations with convolutions. By doing so, it forces the pooling operation to perform dimension reduction along each position of the individual channel, enabling more robust reconstructions of the time series data.

\subsubsection{HeartSpace~\cite{heartspace}} 
HeartSpace designs a deep auto-encoder module to reconstruct the day-specific time series to latent representations, and optimizes parameters through the Siamese-triplet network optimization strategy. The optimization is based on the reconstruction loss and Siamese-triple loss, where the Siamese-triple loss is constructed by sampling the intra-series and the inter-series. To mitigate the side effect caused by zero padding, HeartSpace introduces a dual-stage gating mechanism into the convolutional auto-encoder module to re-weight the hidden units. The channel-wise and temporal-wise gating mechanisms focus on capturing dependencies in the channel dimension and temporal dimension, respectively.

\subsubsection{MBTGMM~\cite{mbtgmm}} 
MBTGMM focuses on learning representations of time series and detecting anomalies based on the reconstruction error, where the anomalies are associated with high reconstruction errors. MBTGMM utilizes a multi-branch transformer-based model, which is designed to capture both short-term and long-term temporal dependencies in the time series while learning the context-aware representations. The learned representations are then fed into a Gaussian mixture model to estimate the density of learned normal representations. By identifying low-density regions, MBTGMM is able to effectively detect anomalies in the raw time series data.

\subsubsection{SimMTM~\cite{SimMTM}}
SimMTM offers an effective approach for recovering masked time points by leveraging the weighted aggregation of neighbors outside the manifold. Directly masking a portion of time points can severely disrupt the temporal variations present in the original time series. Such disruption poses significant challenges for guiding representation learning of time series during reconstruction. SimMTM facilitates the reconstruction by assembling the ruined but complementary temporal variations from multiple masked series. In addition to the reconstruction loss, SimMTM incorporates a constraint loss to guide the series-wise representation learning based on the neighborhood assumption of the time series manifold.

\subsubsection{CRT~\cite{crt}} 
Motivated by observations that dropping is better than masking, adding phase aids in frequency learning, and cross-domain learning is more effective than single-domain learning, CRT utilizes the cross-domain dropping-reconstruction task to facilitate effective representation learning. CRT slices the input into patches and randomly drops some sliced patches, where the remaining patches are projected as 3 types (time, magnitude, or phase) of embeddings using a transformer encoder. Then, CRT uses a transformer decoder to reconstruct original data from the 3 types of data separately, enabling the model to receive segments of real data without corrupted zeros portions. This helps maximally preserve the global context and capture the intrinsic long-term dependencies.

\section{Self-supervised Learning Methods}
\label{s5}

In past years, supervised learning has achieved success in directly learning semantic information from a vast number of labeled samples. However, it requires a large amount of annotated training data and is often tailored to specific tasks, making the trained models less transferable. To address these challenges, self-supervised learning has emerged as a promising technique in representation learning. It eliminates the need for expensive manual labeling by designing diverse pretext tasks that automatically generate useful supervised signals from the original data. In this section, we classify self-supervised learning methods into 3 categories: adversarial methods, predictive methods and contrastive methods. These categories reflect different types of pretext tasks employed in different methods to generate self-supervised signals. Adversarial methods learn representations by distinguishing between the generated fake data and the real data. Predictive methods focus on learning representations by predicting missing or transformed parts of the original input. Contrastive methods instead focus on learning discriminative representations by emphasizing similarities and differences between samples.

\subsection{Adversarial Methods}
Adversarial methods treat distinguishing real data from fake data as a pretext task to learn robust representations for time series~\cite{rdcgan}. In previous studies, Generative Adversarial Networks (GANs) demonstrate their ability on augmenting smaller time series datasets by generating previously unseen data~\cite{survey_ts_gan}. The explicit reconstruction cost minimized by GANs tends to emphasize the higher-level semantic details in learned representations~\cite{DBLP:conf/nips/DonahueS19}. Conventional GAN-based methods facilitate the acquisition of representations by leveraging backpropagation signals through a competitive process that involves two parameterized feed-forward neural networks, namely, a generator and a discriminator~\cite{gan_overview}. These methods usually create a two-player mini-max game, where the generator tries to improve its ability to deceive the discriminator, and the discriminator tries to become more adept at distinguishing real from fake data~\cite{rdcgan}. By iteratively training the generator and discriminator in an adversarial manner, such methods encourage the generator to learn representations that effectively capture important characteristics of the underlying data distribution. Adversarial methods differ from reconstruction-based methods in that they consider not only reconstruction errors from the generator but also prediction errors from the discriminator. This joint optimization process encourages the generator to generate more realistic samples while the discriminator becomes more proficient at detecting generated samples. In addition to conventional GANs, variants of GANs can be designed, such as incorporating RNNs~\cite{rgan} or autoregressive models~\cite{timegan} to capture temporal dependencies. Furthermore, unlike studies~\cite{timegan,madgan,tsgan,rgan} that utilize GANs to generate synthetic time series data, TimeVAE~\cite{timevae} instead employs variational auto-encoders to facilitate data generation. According to~\cite{timevae}, the standard approach of discriminating real data versus synthetic data is insufficient to capture temporal dependencies. Hence, TimeVAE~\cite{timevae} injects temporal structures within the decoder to ensure that the synthetic data exhibit desired temporal patterns. Further details about the aforementioned adversarial methods are provided as follows.

\subsubsection{TimeGAN~\cite{timegan}} 
TimeGAN merges the advantages of the GAN framework's versatility with the control afforded by training in autoregressive models. It achieves this by integrating a learned embedding space that is jointly optimized using both supervised and adversarial objectives. This approach enables the adversarial network to capture and replicate the temporal dynamics present in the training data during the sampling process. The embedding network plays a crucial role by establishing a reversible mapping between the features and latent representations, thereby effectively reducing the dimensionality of the adversarial learning space.

\subsubsection{MAD-GAN~\cite{madgan}} 
MAD-GAN performs multivariate anomaly detection with GAN to identify whether the testing data conform to the normal data distributions. MAD-GAN leverages a generator and discriminator trained by the GAN framework to capture the intricate multivariate correlations contained in time series data. The GAN-trained generator produces fake time series by taking sequences from a random latent space as input, while the GAN-trained discriminator learns to detect fake data from real data in an unsupervised manner. To optimize the model, MAD-GAN designs a combined anomaly score called DR-score, which incorporates both discrimination and reconstruction aspects. This allows for the effective detection of anomalies within the time series data.

\subsubsection{TSGAN~\cite{tsgan}} 
Time Series GAN (TSGAN) is designed to address the challenges of generating realistic time series data, particularly in domains where data acquisition is difficult or limited. TSGAN utilizes 2 GANs, one regular and one conditional, to model and generate synthetic time series examples. The architecture leverages the concepts of Wasserstein GANs (WGANs)~\cite{wgans} and conditional GANs (CGANs)~\cite{cgans} to improve the quality and relevance of the generated data. TSGAN has shown promising results in generating realistic 1D signals across various data types, such as sensor, medical, simulated, and motion data. It also demonstrates superior performance in real applications, such as creating classifiers based on synthetic data.

\subsubsection{RGAN~\cite{rgan}} 
Recurrent GAN (RGAN) generates real-valued multi-dimensional time series with improved realism and coherence. RGAN works by utilizing recurrent neural networks (RNNs) in both the generator and discriminator components of the GAN framework. By incorporating RNNs, RGAN can capture temporal dependencies and generate sequences that exhibit realistic patterns and dynamics. This approach is particularly valuable for generating realistic medical time series data, where capturing temporal relationships is crucial for accurate modeling and analysis. The use of RNNs in both the generator and discriminator sets RGAN apart from traditional GAN architectures and contributes to its effectiveness in generating realistic time series data.

\subsubsection{TimeVAE~\cite{timevae}}
TimeVAE utilizes a decoder that allows for the incorporation of user-defined distributions, enabling flexibility in the generated data. TimeVAE is trained by using the evidence lower bound loss function. A weight is used on the reconstruction error, allowing for the adjustment of emphasis placed on the reconstruction loss compared to the KL-Divergence loss between the encoded latent space distribution and the prior. In addition, TimeVAE injects temporal structures including level, trend, and seasonal components into the data generation process within the decoder to enhance the interpretability of the modeled data generation process.

\subsubsection{DIVERSIFY~\cite{DIVERSIFY}}
DIVERSIFY employs an adversarial game strategy to simultaneously maximize the 'worst-case' distribution scenario while minimizing distribution divergence. Specifically, it learns to segment time series data into distinct latent sub-domains, aiming to maximize the distribution gap at the segment level to preserve diversities. Simultaneously, it also focuses on reducing distribution divergence between these obtained latent domains to achieve domain-invariant representations. This approach leverages the inherent presence of diverse latent distributions in time series data, such as distinct activity patterns exhibited by multiple individuals.

\subsection{Predictive Methods} 
Predictive methods learn meaningful representations that capture the underlying shared information between different parts of time series data by maximizing the mutual information derived from related slices of original time series or diverse views generated by data augmentations. The size of mutual information indicates the strength of dependencies between related slices or augmented views. Compared to reconstruction-based methods, predictive methods help eliminate the need to reconstruct the complete input and learn representations by predicting future, missing or contextual information. Hence, they usually utilize context prediction or cross-view prediction as the pretext task to predict partial data based on limited views. The key insight of predictive methods is to learn representations by predicting future or mixed values of partial time series~\cite{wave2vec,cass,cpc}, predicting whether time windows are sampled from the same temporal context or not~\cite{tst, eegssl}, or predicting the cross-view representations of original samples~\cite{tstcc}. Such mechanism can be served as a pretext task for contrastive learning as well, and methods such as CPC~\cite{cpc} and TS-TCC~\cite{tstcc} combine predictive coding and noise-contrastive estimation to induce the latent space to capture information that is maximally useful for prediction. Further elaboration about CPC~\cite{cpc} and TS-TCC~\cite{tstcc} will be introduced in Section~\ref{lal}. Below are the additional details of the remaining predictive methods used for unsupervised representation learning.
 
\subsubsection{TST~\cite{tst}} 
TST employs a transformer encoder to extract dense representations of time series. This is achieved through an input denoising objective, where the model is trained to reconstruct the entire input under noise corruption, typically using Gaussian noise. Unlike methods that rely on additional unlabeled samples, TST maximizes the utilization of existing samples for representation learning by employing an unsupervised autoregressive objective. TST encourages the model to attend to both preceding and succeeding segments within individual variables, as well as the contemporary values of the other variables within time series, enabling the model to capture complex relationships across different dimensions.

\subsubsection{EEG-SSL~\cite{eegssl}}
EEG-SSL focuses on predicting whether time windows are sampled from the same temporal context or not. EEG-SSL aims to learn informative representations from EEG time series data by employing 2 kinds of temporal-based pretext tasks: relative positioning and temporal shuffling. In the relative positioning task, EEG-SSL predicts whether 2 time windows in the EEG signals are closer or farther apart in time. The temporal shuffling task involves predicting whether the selected time windows are in their original order or have been shuffled. By training on these pretext tasks, EEG-SSL is able to capture temporal dependencies and extract meaningful representations from EEG time series data.

\subsubsection{Wave2vec~\cite{wave2vec}} 
Wave2vec aims to predict the next time step based on the given signal context. Wave2vec consists of 2 multi-layer convolutional neural networks: an encoder network that converts raw audio signals into a latent space, and a context network that combines multiple time-steps of the encoder's output to obtain contextualized representations. The pre-training approach allows Wave2vec to learn meaningful audio representations that can capture the temporal dependencies and facilitate the downstream audio processing tasks.

\subsubsection{CaSS~\cite{cass}}
Prior research primarily emphasizes the pretext task and tends to overlook the intricate issue of encoding time series data, which often yields unsatisfactory outcomes. In contrast, CaSS comprehensively addresses this challenge from 2 aspects: encoder and pretext task. CaSS introduces a channel-aware transformer architecture, which is specifically designed to capture the intricate relationships between different time channels within time series. Furthermore, CaSS integrates 2 innovative pretext tasks, namely next trend prediction and contextual similarity, to further enhance the encoding process.

\begin{table*}[thp]
\setlength{\tabcolsep}{2pt}
\centering
\small
      \begin{tabular}{l|p{7.5cm}|p{1.8cm}|l|p{2.2cm}|p{2cm}|p{0.8cm}}
      \toprule[1.2pt]
  Models &  Main Contributions & Backbones & DA & Datasets & Evaluations & Metrics \\
      \midrule
     SimCLR~\cite{simclr} & SimCLR advocates for a learnable nonlinear transformation bridging the representation and the contrastive loss to enhance the quality of learned representations, and highlights the significance of larger batch sizes, more training steps, and the composition of data augmentations. & ResNet & \checkmark &  ImageNet, CIFAR-10 &  Classification, Transferability & ACC \\ 
      \midrule
    TimeCLR~\cite{timeclr} & TimeCLR extends SimCLR to the time series domain,  amalgamating the benefits of dynamic time warping data augmentation tailored for univariate time series, and the potent feature extraction capability of InceptionTime, thus facilitating the acquisition of representations.  & InceptionTime & \checkmark & Hand Atlas & Classification & ACC, F1\\
     \midrule
   MoCo~\cite{moco} &  MoCo maintains a dynamic queue to enrich the set of negative samples, and proposes a slowly progressing key encoder, implemented as a momentum-based moving average of the query encoder, to preserve the consistency of key representations. & ResNet & \checkmark &  ImageNet, CIFAR-10 &  Classification, Transferability & ACC \\ 
  \midrule
  BYOL~\cite{byol} & BYOL removes the need of using negative samples, and employs a predictor on top of the online network to learn the mapping from the online encoder to the target encoder, which helps prevent mode collapse. & ResNet & \checkmark &  ImageNet, CIFAR, SUN397, VOC07, DTD &  Classification, Transferability & ACC \\ 
      \midrule
  CPC~\cite{cpc} & CPC extracts compact latent representations to encode predictions over the future observations by combining autoregressive modeling and noise-contrastive estimation with intuitions from predictive coding.  &  ResNet, GRU & \multicolumn{1}{c|}{} &  LibriSpeech, ImageNet, BookCorpus & Classification & ACC \\ 
      \midrule
   SimSiam~\cite{simsiam} &  Simsiam learns representations by using Siamese architectures without negative sample pairs, large batches and momentum encoders. SimSiam tackles the mode collapse problem by using the stop-gradient mechanism. & ResNet & \checkmark & ImageNet, VOC07, COCO & Classification, Transferability  & ACC \\ 
      \midrule
    MCL~\cite{mcl} & MCL learns representations through the injection of noise. Inspired by label smoothing, MCL adopts mixup that creates new samples by convex combinations of training examples, and predicts the strength of the mixing component based on 2 data points and the augmented sample. & FCN & \checkmark & UCR Datasets, UEA Datasets & Classification, Transferability & ACC \\ 
   \bottomrule[1.2pt]
   \end{tabular}
   \caption{Summary of instance-level contrastive methods for time series representation learning. DA indicates whether the model utilizes data augmentations. ACC refers to accuracy, F1 refers to F1 score. }
   \label{tab:summary_instance}
\end{table*}

\subsection{Contrastive Methods}
\label{lal}
Contrastive methods learn meaningful representations from time series by optimizing self-discrimination tasks. Instead of directly modeling the complex raw data, they employ pretext tasks that leverage the underlying similarity between samples, which eliminates the need for reconstructing the complete input and allows for the discovery of contextualized underlying factors of variations~\cite{review_ssl_speech}. Contrastive methods typically generate augmented views of the raw data through various transformations and then learn representations by contrasting positive samples against negative samples~\cite{empirical_graph_cl}. Exploring negative samples in a completely unsupervised manner can be challenging for contrastive learning, as it often encounters hard negative samples with features that are highly similar to anchors but have different labels. To address this, researchers have explored various techniques to introduce more negatives, such as increasing the batch size~\cite{simclr} or exploiting external data structures~\cite{moco,pcl}. There are also studies that remove the need of using negative samples and instead use other strategies such as utilizing a predictor~\cite{byol}, a stop-gradient operation~\cite{simsiam}, or clustering~\cite{swav} to learn the mapping. Within the realm of computer vision, contrastive methods are commonly categorized into instance-level and prototype-level methods, with each approach focusing on different levels of contrast. Additionally, there are specific methods tailored for time series data, which consider the temporal-level contrast. We analyze the similarities and differences between these representative contrastive models in terms of their main contributions, backbone architectures, data augmentations (DA), datasets, evaluation strategies, and performance metrics. Contrastive learning enables the discovery of informative representations in time series data without relying on explicit reconstruction tasks. This analysis provides insights into the key characteristics and approaches of representative contrastive models in the field of representation learning for time series data. The concise overview of instance-level, prototype-level, and temporal-level methods can be found in Table~\ref{tab:summary_instance}, Table~\ref{tab:summary_prototype}, and Table~\ref{tab:summary_temporal}, respectively. The following sections provide more in-depth details about these methods.

\subsubsection{Instance-level Contrastive Learning Models}
Instance-level contrastive learning models treat individual samples independently for the purpose of instance discrimination. They utilize data augmentations to transform original inputs into a new embedding space. Within this space, augmentations derived from the same sample are considered as positive pairs, while those from different samples are treated as negative pairs. During training, these models are optimized by maximizing the similarity between representations of positive pairs, while simultaneously minimizing the similarity between representations of negative pairs. Instance-level contrastive methods achieve significant performance improvement by setting larger batch size~\cite{simclr}, using stronger augmentations~\cite{timeclr,mcl}, or introducing an additional storage space to store more contrastive candidates~\cite{instdisc, moco}. However, these methods can be memory-intensive, and ensuring the consistency of representations is challenging when randomly extracting negative samples from an additional storage space. Recent advancements in non-contrastive learning, which eliminate the need for negative samples, have exhibited promising outcomes~\cite{DBLP:conf/aistats/PokleTLR22}. BYOL~\cite{byol} and SimSiam~\cite{simsiam} serve as exemplars, achieving remarkable results without the reliance on negative samples. These methods leverage an additional learnable predictor and employ a stop-gradient operation to prevent collapsing, contributing to their successful performance~\cite{DBLP:conf/icml/TianCG21}. Table~\ref{tab:summary_instance} provides a succinct summary of instance-level contrastive methods for time series representation learning. Further elaboration concerning these methods are explicated below.

\paragraph{SimCLR~\cite{simclr}}
SimCLR greatly improves the quality of representation learning by introducing a learnable nonlinear projection between the representation and the contrastive loss. Since the representation obtained from the encoder retains information related to data augmentations, the role of the nonlinear layer is to remove such information and allow representations to reflect the essence of data. This aids in preventing similarity-based losses from discarding crucial features during training. SimCLR randomly draws a minibatch of $N$ samples and generates $2N$ augmented views by applying geometric and appearance transformations sequentially from the same minibatch. Among these augmented views, a positive pair is formed by selecting $2$ views derived from the same original sample. The remaining $2(N-1)$ augmented views within the minibatch are treated as negative ones. Such a sampling strategy has been widely adopted in various contrastive models.

\paragraph{TimeCLR~\cite{timeclr}}
Direct applying SimCLR to the time series field usually performs poorly due to data augmentation and the feature extractor not being adapted to the temporal dependencies within the time series data. TimeCLR adopts Dynamic Time Warping (DTW) data augmentation not only generates DTW-targeted phase shifts and amplitude changes, but also retains the structure and feature information of the time series. In addition, TimeCLR adopts InceptionTime which has good feature extraction capabilities as the feature extractor, to convert the time series into the corresponding representations in an end-to-end manner.

\paragraph{MoCo~\cite{moco}}
MoCo leverages a moving-averaged momentum encoder and maintains a dynamic queue to enhance negative sampling in contrastive learning. The dynamic queue stores a diverse set of samples, enabling the decoupling of dictionary size from the minibatch size and facilitating the contrastive learning process. In each minibatch, MoCo treats the encoded queries and their corresponding keys as positive sample pairs, while the negative samples are selected from the remaining samples in the dynamic queue. This approach helps enhance the quality and diversity of the contrastive learning process in MoCo.

\paragraph{BYOL~\cite{byol}}
BYOL utilizes a pair of neural networks to perform training, namely the online network and the target network. The online network is trained to predict the representation of the target network, and the target network is updated with a slow-moving average of the online network. BYOL exclusively treats 2 different augmented views of the same sample as the positive pair and learns representations without explicitly using negative pairs, which provides new insight into negative sampling in contrastive learning. Then, the augmentations of the same sample are fed into both the online and target networks, which interact and mutually learn from one another. BYOL directly bootstraps the representations, which makes it more robust to the choice of augmentations.

\paragraph{CPC~\cite{cpc}}
Contrastive Predictive Coding (CPC) encodes the underlying shared information between different parts of the original high-dimensional signal, while discarding low-level information and noise that is more local. The combination of predictive coding and a probabilistic contrastive loss enables CPC to extract slow features that maximize the mutual information between the encoded representations that span many time steps. CPC utilizes an encoder and an autoregressive model to reduce the dimensionality of the high-dimensional signal, and then uses noise-contrastive estimation to predict future representations. The encoder maps the input high-dimensional signal to a latent vector in a lower-dimensional space, and the autoregressive model aggregates the latent vectors from previous and current time steps.

\paragraph{SimSiam~\cite{simsiam}}
Simsiam explores the use of Siamese architectures in representation learning and adopts the stop-gradient operation to prevent collapsing, while getting rid of negative samples, large batches and momentum encoders. The learning mechanism of SimSiam is similar to the Expectation- Maximization (EM) algorithm to estimate the expected value after data augmentation. Siamese networks are chosen as the backbone due to their ability to incorporate inductive biases for modeling invariance, which aids in capturing essential features during representation learning.

\paragraph{MCL~\cite{mcl}}
The inherent invariances within time series data are often unknown beforehand, and incautious application can lead to representations where dissimilar samples are embedded in close proximity. MCL exploits a data augmentation scheme in which new samples are generated by mixing 2 data samples with a mixing component. The pretext task motivated through label smoothing is to predict the strength of the mixing component based on the 2 data points and the augmented sample. In addition, MCL is tasked with predicting the mixing factor instead of hard decisions to tackle the problem of overconfidence in neural networks.

\begin{table*}[thp]
\setlength{\tabcolsep}{2pt}
\centering
\small
      \begin{tabular}{l|p{7.6cm}|p{1.6cm}|l|p{2.3cm}|p{2.2cm}|p{1.3cm}}
      \toprule[1.2pt]
  Models &  Main Contributions & Backbones & DA & Datasets & Evaluations & Metrics \\
      \midrule
  SwAV~\cite{swav} &  SwAV optimizes the learned representations by developing a swapped prediction mechanism to simultaneously perform scalable online clustering while enforcing the consistency between cluster assignments produced for different augmented views of the same sample. & ResNet & \checkmark &  ImageNet, VOC07, COCO &  Classification, Transferability & ACC \\
  \midrule
  PCL~\cite{pcl} &  PCL formulates prototypical contrastive learning as an Expectation-Maximization algorithm to perform clustering and representation learning iteratively. The E-step aims to find the distribution of prototypes via clustering and the M-step aims to optimize the network via contrastive learning.
  & ResNet & \checkmark & ImageNet, VOC07, Places205, COCO & Classification, Clustering, Object Detection & ACC, AMI \\ 
    \midrule
  CCL~\cite{ccl} &  CCL refines the learned representations acquired through deep convolutional neural networks by discovering dataset clusters with high purity and typically few samples per cluster, and leverage these cluster assignments as the potentially noisy supervision. & CNN & \checkmark &  BBT-0101, BF-0502, ACCIO & Classification, Clustering & ACC, Precision, Recall, F1 \\ 
      \midrule
  SCCL~\cite{sccl} & SCCL leverages contrastive learning for short text clustering to promote better separated and less dispersed clusters. It effectively combines the top-down clustering with the bottom-up instance-wise contrastive learning to achieve better inter-cluster distance and intra-cluster distance. & DistilBERT & \checkmark & AgNews, Tweet, SearchSnippets, StackOverflow, Biomedical, Googlenews & Clustering & ACC, NMI\\
  \midrule
  CC~\cite{cc} & CC seamlessly integrates deep clustering and representation learning by revealing that the row and column of the feature matrix intrinsically correspond to the instance and cluster representation when projecting instances into a subspace whose dimensionality is equal to the cluster number.  &  ResNet & \checkmark & CIFAR-10, CIFAR-100, STL-10, ImageNet-10  & Classification, Clustering  & ACC, NMI, ARI \\ 
      \midrule
  SLIC~\cite{slic} & SLIC combines iterative clustering with multi-view encoding and temporal discrimination to learn view-invariant video representations and fine-grained motion features. The clustering assignments are used to guide the sampling of positive and negative pairs for updating representations. & CNN & \checkmark & UCF101, HMDB51, Kinetics400 & Classification, Video Retrieval & Recall\\
  \midrule
  MHCCL~\cite{mhccl} & MHCCL incorporates the implicit semantic information obtained from hierarchical clustering to guide the construction of contrastive pairs. MHCCL exploits downward masking to filter out fake negatives and supplement positives, while also employing upward masking to refine prototypes. & ResNet & \checkmark & HAR, WISDM, SHAR, Epilepsy, UEA Datasets &  Classification & ACC, MF1, $\kappa$ \\ 
      \bottomrule[1.2pt]
   \end{tabular}
   \caption{Summary of prototype-level contrastive methods for time series representation learning. DA indicates whether the model utilizes data augmentations. ACC refers to accuracy, AMI refers to adjusted mutual information, NMI refers to normalized mutual information, F1 refers to F1 score, ARI refers to Adjusted Rand Index, MF1 refers to macro-averaged F1 score, and $\kappa$ refers to cohen's kappa coefficient.}
   \label{tab:summary_prototype}
\end{table*}

\subsubsection{Prototype-level Contrastive Learning Models}
\label{prototype-level cl}
To address the limitation that instance-level contrastive learning models tend to treat semantically similar samples as negatives, prototype-level contrastive learning models break the independence between samples, and explore to exploit the implicit semantics shared by samples within the same cluster. By incorporating the additional prior information brought by clustering, the learned representations are expected to preserve more higher-level semantic information~\cite{mhccl}. Prototype-level contrastive learning models consider groups of similar samples as clusters and pull the representations of samples from different augmented views but belonging to the same class (with similar semantics) together in the embedding space during contrastive learning. For instance, CC~\cite{cc}, SwAV~\cite{swav} and PCL~\cite{pcl} are such methods that perform clustering-based discrimination between groups of similar samples. However, these methods require prior knowledge to pre-specify the number of clusters, which is non-trivial for the unlabeled time series data. What's worse, the flat clustering algorithms they adopt can only capture a single hierarchy of semantics, which is insufficient to reflect the ground-truth data distributions. Recently, methods such as CCL~\cite{ccl}, SLIC~\cite{slic} and MHCCL~\cite{mhccl} have merged to incorporate hierarchical clustering into contrastive learning. By exploring the hierarchical structures, these methods are able to capture diverse granularities of semantics in time series data. The differences among the varying prototype-level contrastive learning models lie in the selection of cluster centroids, clustering methods, and cluster-level sampling strategies. Table~\ref{tab:summary_prototype} provides a succinct summary of prototype-level contrastive methods for time series representation learning. Further elaboration on these methods is explicated below.

\paragraph{SwAV~\cite{swav}}
SwAV enhances representation learning in a distinctive manner by leveraging cluster assignments and a swapped prediction mechanism to perform contrasting. SwAV assigns samples to different clusters rather than directly contrasting samples, where the cluster assignments are used to construct the implied contrastive pairs. Instead of approximating each sample into a hard assignment, the soft assignment is produced by the Sinkhorn-Knopp algorithm. Then, the implied positive pair is formed by the one augmented view and the cluster assignment of the other view, and the remaining pairs are treated as implied negative pairs. The swapped prediction mechanism involves predicting the cluster assignment of one augmented view from the other augmented view.

\paragraph{PCL~\cite{pcl}}
PCL formulates the prototypical contrastive learning framework as an Expectation-Maximization (EM) algorithm. The framework consists of 2 iterative steps: the E-step and the M-step. In the E-step, PCL performs clustering to find the distribution of prototypes. In the M-step, the network is optimized through contrastive learning. The total loss is composed of a standard InfoNCE loss for instance-level contrast and a ProtoNCE loss for prototype-level contrast. The instance-level sampling follows the same strategy as MoCo~\cite{moco}. For prototype-level contrast, each sample forms a positive pair with its corresponding prototype, which represents the centroid of the cluster it belongs to. The remaining prototypes are considered negative ones.

\paragraph{CCL~\cite{ccl}}
Clustering-based Contrastive Learning (CCL) learns cluster-based representations without relying on prior knowledge of the number of clusters. It automatically obtains labels from natural groupings using hierarchical clustering that does not require any hyper-parameters. To construct positive and negative pairs, CCL leverages clusters with high purity and a few samples at early hierarchical partitions. For each data point within a cluster, CCL randomly samples from the same or nearest cluster to create one positive pair, while using samples from the farthest clusters to create 2 negative pairs. This approach ensures that the generated pairs capture both intra-cluster similarities and inter-cluster differences.

\paragraph{SCCL~\cite{sccl}}
SCCL focuses on short text data and develops a joint model that leverages the beneficial properties of instance-wise contrastive learning to improve unsupervised clustering. This is implemented by jointly optimizing a top-down clustering loss over the original data samples and a bottom-up instance-wise contrastive loss over the associated augmented pairs. The Student's t-distribution is utilized to compute the probability of assigning each sample to different clusters. The instance-wise contrastive learning focuses on distance discrimination, while unsupervised clustering focuses on clustering-based discrimination between groups of samples with higher-level semantic concepts.

\paragraph{CC~\cite{cc}}
Contrastive Clustering (CC) regards rows of the feature matrix projected from augmentations as soft labels of samples, while treating columns as prototype-level representations distributed over the dataset. This is motivated by the observation of ``label as representation" when the dimensionality of representations during projection matches the number of clusters. CC simultaneously learns discriminative representations and performs online clustering in an end-to-end manner. It adopts a dual contrastive learning strategy that operates at both the instance level and cluster level to extract informative representations.

\paragraph{SLIC~\cite{slic}}
SLIC is an iterative clustering-based contrastive learning framework to learn feature representations from human action videos. SLIC integrates iterative clustering with multi-view encoding and a temporal discrimination loss to sample harder positives and negatives during pre-training. The encoder is optimized by incorporating a temporal discrimination loss and an instance-based triplet loss. To be specific, SLIC alternates between periodically clustering video representations to produce cluster assignments, which are used to inform the sampling of positive and negative pairs to update the video representations, by minimizing a triplet margin loss.

\paragraph{MHCCL~\cite{mhccl}}
MHCCL incorporates implicit hierarchical semantic information obtained by hierarchical clustering into contrastive learning. This is inspired by the observation of the multi-granularity of clustering. The novel bidirectional masking strategies are employed to guide the selection of positive and negative pairs for multi-level contrast. Specifically, upward masking helps remove outliers while refining prototypes to accelerate the hierarchical clustering process and enhance the quality of clustering. Meanwhile, downward masking helps supplement latent positives and filter out fake negatives for effective contrastive learning. In this way, MHCCL enables the extraction of informative representations that capture both fine-grained and high-level semantic information.

 \begin{table*}[thp]
\setlength{\tabcolsep}{2pt}
\centering
\small
      \begin{tabular}{l|p{7.5cm}|p{1.8cm}|l|p{2.3cm}|p{2cm}|p{1.5cm}}
      \toprule[1.2pt]
  Models &  Main Contributions & Backbones & DA & Datasets & Evaluations & Metrics \\
    \midrule
  TS2Vec~\cite{ts2vec} &  TS2Vec utilizes multi-scale contextual information at both timestamp-level and instance-level to distinguish positive and negative samples, thereby improving the generalization capability of the representation model and effectively handling time series data with missing values. & Dilated CNN & \checkmark & UEA Datasets, ETT Datasets, Electricity & Classification, Forecasting & ACC, MSE, MAE\\ 
      \midrule
   TS-TCC~\cite{tstcc} & TS-TCC constructs simple yet efficient time-series-specific augmented views to perform temporal and contextual contrasting, and designs a tough cross-view prediction task to learn the robust and discriminative representations. &  CNN, Transformer& \checkmark &  HAR, Sleep-EDF, Epilepsy, FD &  Classification, Transferability & ACC, MF1 \\ 
  \midrule
    TNC~\cite{tnc} &  TNC learns the underlying dynamics of non-stationary signals and models the progression over time by defining a temporal neighborhood. It incorporates concepts from Positive-Unlabeled learning to account for potential bias introduced in sampling negative examples for contrastive loss.
    &  Bidirectional RNN &  &  Simulation, ECG Waveform, HAR & Classification, Clustering &  ACC, AUPRC, Silhouette Score, DBI \\
   \midrule
    TCL~\cite{tcl} &  TCL learns representations for time series that allow optimal discrimination of different time segments based on the temporal non-stationary structure captured by nonlinear independent component analysis. & Nonlinear ICA &  & MEG   &  Classification & ACC \\ 
    \midrule
   T-Loss~\cite{tloss} & T-Loss learns scalable representations by taking highly variable lengths and sparse labeling properties of time series data into account. It employs an efficient triplet loss with time-based negative sampling to differentiate anchors from negatives, and assimilate anchors with positives.& Causal CNN &  &  UEA Datasets, UCI Datasets, UCR Datasets &  Classification, Transferability & ACC \\
  \midrule
   BTSF~\cite{btsf} & BTSF applies dropout to generate diverse views for representation learning, and devises iterative bilinear temporal-spectral fusion to explicitly model pairwise cross-domain dependencies for discriminating and enriching representations in a fusion-and-squeeze manner. & Causal CNN & \checkmark &  HAR, Sleep-EDF, ECG Waveform, ETT Datasets, Weather, SAaT, WADI, SMD, SMAP, MSL & Classification, Forecasting, Anomaly Detection & ACC, AUPRC, MSE, MAE, F1 \\
  \midrule
  CoST~\cite{cost} &  CoST simulates interventions on the error variable via data augmentation and exploits prior knowledge to learn time series representations. It leverages inductive biases in the model architecture to learn disentangled seasonal and trend representations via contrastive learning. & Causal CNN & \checkmark & ETT Datasets, Electricity, Weather & Forecasting & MSE, MAE\\
   \bottomrule[1.2pt]
   \end{tabular}
   \caption{Summary of temporal-level contrastive methods for time series representation learning. DA indicates whether the model utilizes data augmentations. ACC refers to accuracy, AUPRC refers to the area under the precision-recall curve, DBI refers to Davies-Bouldin Index, F1 refers to F1 score, MF1 refers to macro-averaged F1 score, MSE refers to mean square error, and MAE refers to mean absolute error.}
   \label{tab:summary_temporal}
\end{table*}

\subsubsection{Temporal-level Contrastive Learning Models}

Although contrastive learning methods have demonstrated remarkable success in the field of computer vision, their application to time series data often ignores the intricate characteristics that are inherent to such data. Acknowledging this limitation, researchers have recently turned their attention towards investigating the influence of timestamps, and have begun developing specifically tailored strategies to address the challenge posed by time series data. Instance-level contrastive learning models are able to learn general representations that capture the overall characteristics of the entire time series. These representations provide a holistic view of data and can be useful for tasks that require a global understanding of the sequence. Temporal-level contrastive learning models instead focus on capturing scale-invariant representations at each individual timestamp. By considering both instance-level and temporal-level representation learning strategies, researchers aim to enhance the capability of contrastive learning methods in capturing the complexities inherent in time series data. Existing temporal-level contrastive learning models either consider the temporal dependencies by leveraging temporal contrasting modules~\cite{tstcc, tloss, tnc}, or focus on capturing multi-scale contextual information across different granularities~\cite{ts2vec,tcl,btsf,cost}. Approaches such as T-Loss~\cite{tloss} and TNC~\cite{tnc} utilize the information from the neighborhood to construct positive and negative samples for contrastive learning. TS-TCC~\cite{tstcc} instead designs time-series-specific augmented views for pair constructions and performs cross-view temporal contrasting to learn representations. There also exist methods that explicitly emphasize the timestamp-level~\cite{ts2vec} or segment-level~\cite{tcl, btsf,cost} representations in addition to the entire time-series level. These models are designed to capture fine-grained temporal patterns and variations, enabling them to extract more detailed and localized information from time series. However, they may ignore the higher-level semantic information which is involved in the entire set of time series. Incorporating potential semantic information such as class labels can help eliminate fake negative samples of the same class, thereby reducing noise in contrastive learning. Table~\ref{tab:summary_temporal} provides a succinct summary of temporal-level contrastive methods for time series representation learning. Further elaboration on these methods is explicated below.

\paragraph{TS2Vec~\cite{ts2vec}}
TS2Vec utilizes multi-scale contextual information with granularities to distinguish samples. It enables the learning of timestamp-level representations, while also supporting instance-level representations of the entire time series by applying max pooling strategy across timestamps. TS2Vec eliminates the need of the input projection layer so that masks can be directly applied to raw input by setting masked timestamps to zeros. Through the masking of latent vectors rather than the raw values, the model acquires the capability to distinguish between masking tokens and original values.

\paragraph{TS-TCC~\cite{tstcc}}
TS-TCC employs a combination of time-series-specific weak augmentations (\emph{i.e.,} jitter-and-scale) and strong augmentations (\emph{i.e.,} permutation-and-jitter) on original data. This approach aims to enhance the quality of the learned features through contrastive learning. TS-TCC designs cross-view temporal and contextual contrasting modules to learn robust and discriminative representations. The temporal contrasting module focuses on capturing the temporal dependency within the data. The contextual contrasting module performs a cross-view prediction task to predict the future of one view based on the past of the other view.

\paragraph{TNC~\cite{tnc}}
TNC addresses potential bias in negative samples by employing a sample weight adjustment strategy. Instead of randomly selecting negative examples from the data distribution, TNC utilizes Positive-Unlabeled (PU) learning without explicitly using negative samples. In PU learning, the combinations of positive and negative samples are treated as unlabeled data. TNC considers samples from the neighborhood as positive examples and samples outside the neighborhood as unlabeled examples. This approach helps mitigate bias and improves the accuracy of the learning process.

\paragraph{TCL~\cite{tcl}}
TCL aims to learn representations that capture the temporal non-stationary structure of time series data, enabling optimal discrimination of different time segments. TCL combines a heuristic principle for analyzing temporal structure with a rigorous treatment of a nonlinear independent component analysis model~\cite{DBLP:conf/icml/ZimmermannSSBB21}. The model is trained through a multinomial logistic regression classifier, which aims to accurately discriminate all time segments in a time series by utilizing the segment indices as labels for the data points.

\paragraph{T-Loss~\cite{tloss}}
T-Loss learns scalable general-purpose representations by considering inherent characteristics of time series, including highly variable lengths and sparse labeling. It introduces an efficient unsupervised triplet loss that utilizes time-based negative sampling and leverages the encoder's resilience to handle time series of unequal lengths. A random sub-series from a given time series is considered the anchor, while any sub-series of the anchor is treated as the positive sample. Sub-series from randomly selected series are treated as negative samples. The model is optimized to distinguish anchors from negatives while assimilating anchors and positives, enabling the learning of discriminative representations.

\paragraph{BTSF~\cite{btsf}}
Prior works primarily use time-based augmentations to sample positive and negative pairs for contrastive training. However, these approaches mainly rely on segment-level augmentations derived from time slicing, which can introduce sampling bias and incorrect optimization due to the loss of global context. Furthermore, they often overlook the incorporation of spectral information in representations. BTSF devises a novel iterative bilinear temporal-spectral fusion approach to explicitly encode the affinities of numerous time-frequency pairs, and employs instance-level augmentation with a simple dropout applied to the entire time series, thereby capturing long-term dependencies more effectively.

\paragraph{CoST~\cite{cost}}
An effective representation should possess the capability to disentangle multiple explanatory sources, enabling robustness against complex and richly structured variations. CoST leverages inductive biases within the model architecture to learn disentangled seasonal and trend representations. Additionally, it introduces a novel frequency domain contrastive loss to encourage the development of discriminative seasonal representations. By providing insights from a causal perspective, CoST highlights the advantages of learning disentangled seasonal-trend representations through contrastive learning for time series forecasting.

\section{Conclusion and Future Directions}
\label{s6}

\subsection{Summary of The Review}
Time series representation learning aims to map the time series into a latent space where the learned semantic-rich representations then can be effectively utilized to benefit downstream tasks such as classification and forecasting. In recent years, numerous unsupervised representation learning methods have been proposed, resulting in a dilemma over selecting the most suitable approach for downstream tasks of interest, especially those tasks on time series data. To fill this gap, in this work, we conduct a thorough literature review of existing methods for time series representation learning, aiming to propose a comprehensive categorization of these approaches. In addition to analysis, we perform empirical evaluations, with a focus on state-of-the-art contrastive methods, using 9 public real-world datasets from diverse domains. The purpose of this experimental evaluation is to offer readers a better, faster, and more comprehensive understanding of cutting-edge research in unsupervised time series representation learning.

\subsection{Future Research Directions}

Contrastive learning has emerged as a powerful unsupervised learning approach due to its capability for capturing rich representations without relying on explicit labels. However, there exist notable challenges that need to be addressed to fully harness the potential of contrastive learning. One critical challenge that warrants consideration is automatic data augmentation, which plays a crucial role in contrastive learning as it generates diverse data covering unexplored input space while preserving correct labels for learning more robust and generalizable representations. However, the augmentation transformations selected by current manual selection strategies based on empirical observations or domain knowledge can be sub-optimal. Hence, developing efficient strategies for automatic data augmentation optimization is imperative. In addition, the primary goal of contrastive learning is to distinguish between similar and dissimilar samples, promoting the learning of discriminative features. Consequently, constructing effective contrastive pairs becomes crucial, particularly in selecting appropriate negative sampling methods, as they directly influence the quality and informativeness of the learned representations. Moreover, efficiency and scalability pose significant challenges for contrastive learning models, especially as the amount and dimensions of input data expand. Developing scalable architectures and efficient algorithms is essential to handle the computational demands of large-scale real-world tasks. Furthermore, overfitting and poor generalization are also common challenges encountered in contrastive learning models. Pre-trained models often struggle to generalize to new data distributions with different statistical properties from the training data, which can hinder their performance on samples from novel classes or unseen domains. By addressing these challenges, the potential of contrastive learning can be fully realized, enabling the extraction of rich representations from unlabeled data, and facilitating breakthroughs in diverse applications. we delve deeper into the aforementioned challenges in the following. 

\subsubsection{Automatic Data Augmentation Optimization}
Data augmentation is a vital component in contrastive learning as it increases diversity and mitigates the risk of false negatives. A theoretical framework~\cite{DBLP:conf/iclr/ZhaoD0Y023} has been proposed to assess the transferability of contrastive learning, with a focus on investigating the impact of data augmentation on its performance. The findings indicate that the effectiveness of contrastive learning in downstream tasks is heavily influenced by the selection of data augmentation techniques. The selection of appropriate augmentations is critical to ensure the preservation of semantic consistency among labels and to foster generalization and robustness, thereby preventing overfitting on the original data. Automating the process of augmentation optimization becomes imperative in order to efficiently identify the most effective augmentation transformations that maximize the learning potential of the model while maintaining the desired properties of the representations~\cite{DBLP:conf/iclr/LingJLJZ23}. The future direction involves developing novel augmentation transformations tailored to time series data and devising efficient strategies to automatically optimize the combinations and parameters of augmentation transformations within a fixed transformation space. Existing data augmentation transformations for time series data~\cite{tstcc,ts2vec,survey_da_dl,survey_da_clf} are heuristic and are usually sub-optimal as they are manually designed or selected based on empirical observations or domain knowledge~\cite{DBLP:conf/icml/NonnenmacherOSR22}. There are several existing works that focus on augmentations specifically designed for image data. For instance, Mixup-noise is utilized in ~\cite{DBLP:conf/icml/VermaLKPL21} to generate augmentations by mixing data samples either at the input or hidden-state levels, without explicitly leveraging the underlying data manifold structure. Distribution divergence is employed in ~\cite{DBLP:journals/pami/WangQ23} between weakly and strongly augmented samples over the representation bank to supervise the retrieval of strongly augmented queries from a pool of instances. Time series data possess unique characteristics and temporal dependencies that require specialized transformations. Conventional time series data augmentations may disrupt the inherent strong temporal dependencies present in the data~\cite{timeclr}. In this case, designing novel augmentation transformations that can capture temporal patterns, variations, and dependencies in time series data is still an open problem that requires further exploration. In addition, optimizing the combinations and parameters of various augmentation transformations becomes challenging due to the large transformation space involved. Every augmentation promotes invariance to a specific transformation, which can be advantageous in certain scenarios and detrimental in others~\cite{DBLP:conf/iclr/Xiao0ED21}. Reinforcement Learning (RL) emerges as a promising solution to automatically search for optimal sub-policies of augmentation transformations and their associated parameters. RL-based methods~\cite{autoaugment, fastaugment} formulate the optimization as a discrete search problem and use reward signals to guide the search process. However, these approaches often demand extensive computational resources due to the size of the search space, making them inefficient for practical usage. Thus, it is still an ongoing problem to design more efficient strategies for automatic data augmentation optimization.

\subsubsection{Designing Appropriate Contrasting Views}
The selection of appropriate augmentation transformations to form the positive pairs is a decisive aspect in contrastive learning~\cite{DBLP:conf/icml/ZimmermannSSBB21}. Designing appropriate contrasting views usually requires domain knowledge, intuition, trial-and-error and luck~\cite{DBLP:conf/cvpr/ChuangHWVJ0JS22}. Existing studies~\cite{DBLP:conf/cvpr/0001MV21,DBLP:conf/nips/Tian0PKSI20} suggest that contrastive learning tends to generate sub-optimal representations in the presence of noisy views such as false positive pairs. This is because representations of these noisy views are forced to align with each other even if there is no apparent shared information. However, there is no guarantee that all task-relevant information is shared between views~\cite{DBLP:conf/cvpr/WangGD022}. Current contrastive learning approaches~\cite{simclr,simclr_2,moco,pcl} commonly consider either the remaining data within the same batch or a large buffer as the default source of negative samples. However, samples with similar higher-level implicit semantics, are easily treated as negative ones~\cite{ccl, slic, mhccl}. Recent studies has started to tackle this problem by incorporating the information of nearest-neighbors to enrich the set of semantic positives as they find that a generic query sample and its neighbors are likely to belong to the same class~\cite{DBLP:conf/iclr/BosnjakRTSWHBPB23, DBLP:conf/iclr/GeWTCS023, DBLP:conf/cvpr/ZhongFRL0S21}. Note that, however, such binary instance labeling is insufficient to measure correlations between different samples, there still exist numerous promising avenues for exploring different approaches to measure the similarity of neighbors, offering greater semantic variations compared to pre-defined transformations~\cite{DBLP:conf/iccv/DwibediATSZ21}. Negative sampling is also critical in constructing contrastive pairs, yet choosing the appropriate sampling method and number of negative samples can be challenging. 
The presence of a collision-coverage trade-off indicates that the ideal number of negative examples should be scaled according to the number of underlying concepts present in the data~\cite{DBLP:conf/aistats/AshGKM22,DBLP:conf/icml/AwasthiDK22}. The occurrence of false negatives adversely affects the discrimination of individual samples as they are erroneously pushed away~\cite{mhccl}. One direction of exploration is to eliminate the need of negative samples~\cite{byol}. However, negative cancellation relies on a sufficient number of clearly distinguishable positive samples. If the positive samples lack clear separability or exhibit significant intra-class variations, the model may still encounter difficulties in discerning between positive and negative pairs, leading to sub-optimal performance. An alternative direction to enhance negative sampling is to employ filtering techniques. These techniques involve selecting negative samples that meet specific criteria, such as similarity or density thresholds, to help the model focus on more informative and diverse negatives. However, it is important to ensure that valuable negative samples are not inadvertently excluded, and that the selected negatives effectively challenge the discrimination ability of models. In addition, clustering techniques can also be employed to group similar negative samples together, reducing the chance of false negatives and promoting more effective discrimination~\cite{mhccl,sccl,ccl}.

\subsubsection{Efficiency and Scalability} 
Due to the inherent characteristics of time series data including variable duration, high dimensionality and high frequency, a time series representation learning framework should satisfy 2 essential criteria simultaneously: efficiency and scalability~\cite{DBLP:journals/corr/abs-2306-06579, DBLP:conf/iclr/Shi0LR0LJ23}. Although researchers have made notable efforts in improving the efficiency and scalability of contrastive models such as adopting early stopping, increasing batch size, applying diverse data augmentations and adopting contrastive regularizations~\cite{simclr_2,moco,byol,DBLP:conf/iclr/Shi0LR0LJ23}, there are still unexplored directions. One potential direction is to develop memory-efficient network architectures to reduce the computational and memory requirements of contrastive models, such as those equipped with weight sharing or pruning. Another way is to incorporate transfer learning, where pre-trained models serve as an initial stage for contrastive learning. Such approaches leverage the knowledge gained from a pre-trained model and fine-tune the model with contrastive learning, thereby reducing training time and computational resources. Additionally, distillation techniques, which compress large models into smaller ones while preserving functionality, are also worth exploring. These approaches have the potential to reduce training time and memory requirements in contrastive learning while maintaining or enhancing performance. Improving the scalability of contrastive learning models is crucial particularly when encountered with large-scale and high-dimensional time series inputs. Techniques such as distributed training, approximate nearest neighbor search and online learning can also be incorporated to enhance the scalability of contrastive models while ensuring efficiency.

\subsubsection{Overfitting and Generalization}
Overfitting and poor generalization are common problems in contrastive learning methods. Recent findings suggest that dropout serves as a form of minimal data augmentation within the contrastive learning process, and removing it could result in a representation collapse in natural language processing tasks~\cite{DBLP:conf/emnlp/GaoYC21}. A theoretical investigation into the generalization ability of contrastive learning reveals key insights regarding its effectiveness, highlighting the significance of the alignment of positive samples, the divergence of class centers, and the concentration of augmented data~\cite{DBLP:conf/iclr/0001YZJ23}. Contrastive learning models may struggle to generalize to new data distributions with different statistical properties from those of the training data. As a result, the pre-trained model becomes proficient at recognizing samples from the base class but performs poorly on samples from novel classes due to overfitting~\cite{fewcon}. In order to address the challenges, a promising research direction is to investigate the application of pre-trained contrastive learning models under the few-shot learning setting. Few-shot learning is a learning paradigm that aims to recognize new samples using only a limited number of labeled examples. While such solutions have been explored extensively in the field of computer vision~\cite{fewshot1,fewshot2}, their applicability in time series data remains an open area for further investigation. Alternatively, another research direction is to leverage few-shot learning to pre-train a model using a small number of labeled examples, enabling the model to capture essential information from limited data. The pre-trained model can then be fine-tuned using contrastive learning, with the objective of maximizing similarity between positive examples and minimizing similarity between negative examples. This two-step process offers the potential for the model to learn representations that are both discriminative and transferable, leading to improved generalization on unseen data and novel classes.

\subsubsection{Robustness Verification}
An effective representation should possess the ability to disentangle diverse explanatory sources, enabling robustness to withstand complex and intricately structured variations~\cite{cost}. The goal of robustness verification is to prove that small perturbations on the designed model do not change the advisories produced for certain inputs~\cite{DBLP:conf/cav/KatzBDJK17}. Recent research has investigated the feasibility of learning provably robust deep neural networks that are verifiably guaranteed to be robust to adversarial perturbations under some specified attack models~\cite{DBLP:conf/nips/WongSMK18}. Although there are methods for discovering the adversarial perturbations, the capacity to verify their absence remains limited~\cite{DBLP:conf/cav/KatzBDJK17}. Existing studies demonstrate that the robustness of contrastive learning models can be improved by augmenting the training set with adversarial samples~\cite{DBLP:conf/icml/Wang022, DBLP:conf/iclr/ZhaoD0Y023, DBLP:conf/cvpr/ChuangHWVJ0JS22}. More future work can be explored on measuring the performance under adversarial attacks or other relevant factors. There also exist studies~\cite{DBLP:conf/icml/XueWM22} which prove that the representation matrix learned by contrastive learning boosts robustness by preventing deep neural networks from overfitting. The robustness metrics used in existing methods such as robust accuracy, however, demonstrate a strong correlation with the attack algorithms, image labels, and downstream tasks in the field of computer vision~\cite{DBLP:conf/icml/Wang022}. These factors hence may introduce inconsistencies and reduce the reliability of robustness metrics in the context of contrastive learning. From another perspective, there are unique characteristics such as frequency and amplitude in time series data that are valuable for consideration. Consequently, it is crucial to explore the design of robustness verification metrics that are specifically tailored to address the intricacies of time series data, ensuring their effectiveness in assessing the robustness of such data.


\input{main.bbl}

\begin{IEEEbiography}[{\includegraphics[width=1in,clip,keepaspectratio]{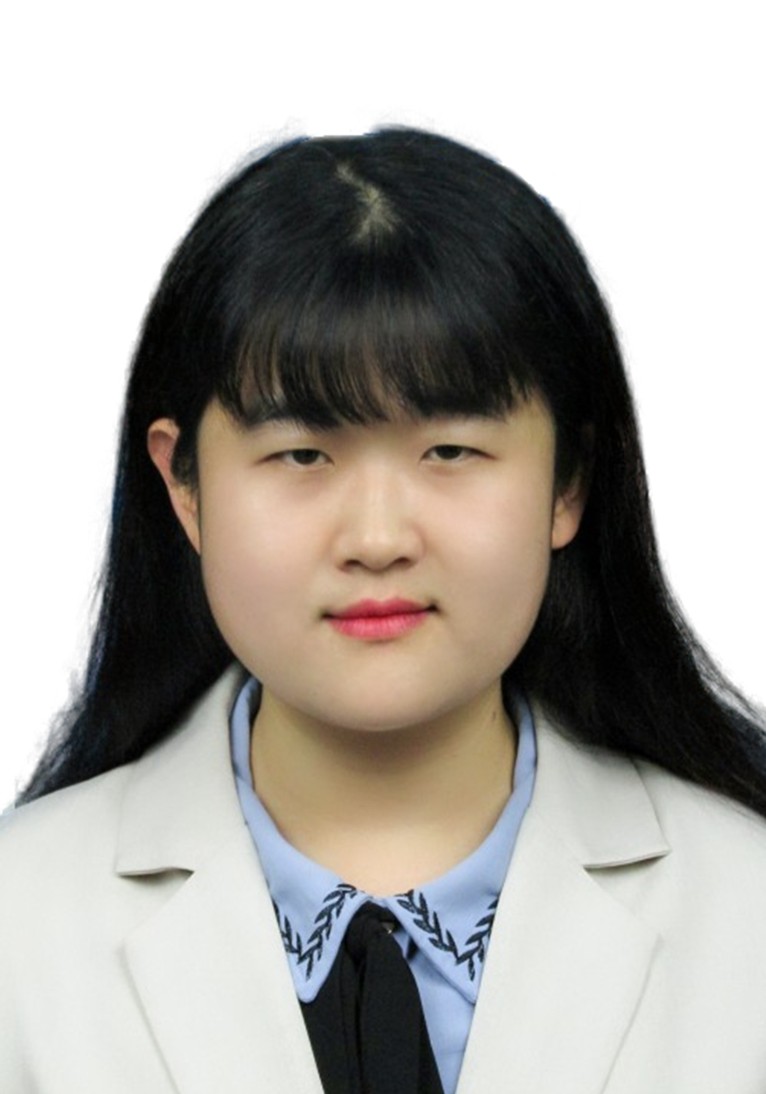}}]{Qianwen Meng} is currently a Ph.D student in School of Software, Shandong University, China, and Joint SDU-NTU Centre for Artificial Intelligence Research,
Shandong University, China. She received the Master’s degree in School of Software, Shandong University, China in 2020, and the Bachelor’s degree in School of Computer Science and Technology, Shandong University, China in 2017. Her research interests mainly lie in representation learning, unsupervised learning, time series modeling, and disease risk prediction.
\end{IEEEbiography}
\vspace{-3em}

\begin{IEEEbiography}[{\includegraphics[width=1in,clip,keepaspectratio]{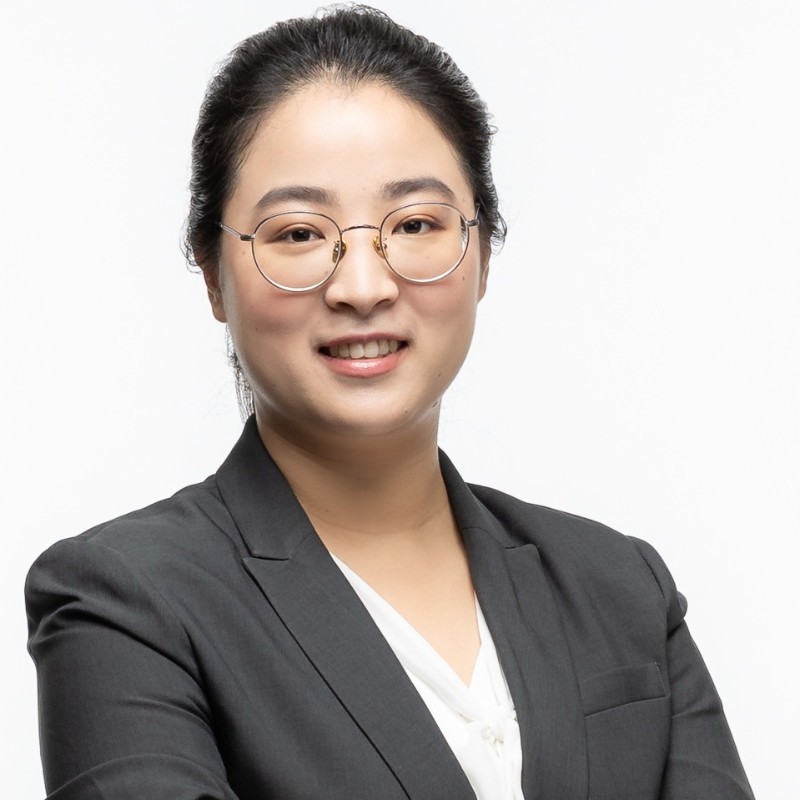}}]{Hangwei Qian} is a research scientist at Centre for Frontier AI Research (CFAR), A*STAR, Singapore. She was previously a Wallenberg-NTU Presidential Postdoctoral Fellow from 2020 to 2022. She obtained her Ph.D. in School of Computer Science and Engineering at NTU, Singapore in 2020 and B.E. from University of Science and Technology of China (USTC) in 2015. Her research interests include unsupervised learning, transfer learning, kernel methods, and wearable-based studies. She has published top-tier academic articles in KDD, AAAI, IJCAI, AIJ and FAccT.
\end{IEEEbiography}
\vspace{-3em}

\begin{IEEEbiography}[{\includegraphics[width=1in,clip,keepaspectratio]{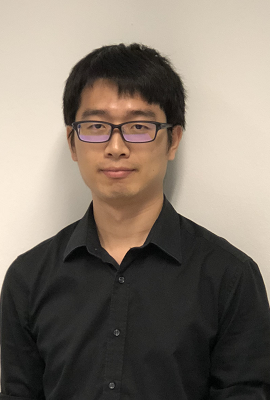}}]{Yong Liu} is a Senior Principal Researcher at Huawei Noah's Ark Lab. Prior to joining Huawei, he was a Senior Research Scientist at Nanyang Technological University (NTU), a Data Scientist at NTUC Enterprise, and a Research Scientist at Institute for Infocomm Research (I2R), A*STAR, Singapore. He received his Ph.D. degree in Computer Engineering from NTU in 2016 and B.S. degree in Electronic Science and Technology from University of Science and Technology of China (USTC) in 2008. His research interests include recommendation systems, natural language processing, and knowledge graph. He has been invited as a PC member of major conferences such as KDD, SIGIR, ACL, IJCAI, AAAI, and reviewer for IEEE/ACM transactions.
\end{IEEEbiography}
\vspace{-3em}

\begin{IEEEbiography}[{\includegraphics[width=1in,clip,keepaspectratio]{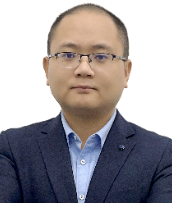}}]{Yonghui Xu}
is a professor at Joint SDU-NTU Centre for Artificial Intelligence Research (C-FAIR), Shandong University. Before that, he was a research fellow in the Joint NTU-UBC Research Centre of Excellence in Active Living for the Elderly (LILY), Nanyang Technological University, Singapore. He received his Ph.D. from the School of Computer Science and Engineering at South China University of Technology in 2017 and BS from the Department of Mathematics and Information Science Engineering at Henan University of China in 2011. His research areas include various topics in Trustworthy AI, knowledge graphs, expert systems and their applications in e-commerce and healthcare. 
\end{IEEEbiography}
\vspace{-3em}

\begin{IEEEbiography}[{\includegraphics[width=1in,clip,keepaspectratio]{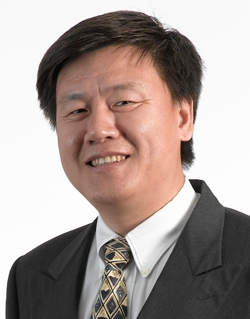}}]{Zhiqi Shen} received the B.Sc. degree in computer science and technology from Peking University, Beijing, China, the M.Eng. degree in computer engineering from the Beijing University of Technology, Beijing, and the Ph.D. degree from Nanyang Technological University, Singapore. He is a Senior Lecturer and a Senior Research Scientist with the School of Computer Science and Engineering, Nanyang Technological University. His current research interests include multiagent systems, goal-oriented modeling, agent augmented interactive media, and interactive storytelling.
\end{IEEEbiography}
\vspace{-3em}  

\begin{IEEEbiography}[{\includegraphics[width=1in,clip,keepaspectratio]{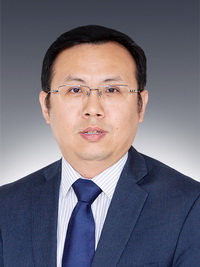}}]{Lizhen Cui} (IET Fellow, IEEE Senior Member) is a professor and doctoral supervisor at School of Software, Shandong University. He is the Co-Director of Joint SDU-NTU Centre for Artificial Intelligence Research (C-FAIR) and Research Center of Software and Data Engineering, Shandong University. He received his B.Sc., M.Sc., and Ph.D. degrees from Shandong University, Jinan, China, in 1999, 2002 and 2005, respectively. He has published over 200 high-level academic articles in TKDE, TPAMI, TPDS, TSC, TCC, TNNLS, AAAI, SIGIR, CIKM, BIBM, ICDCS, DASFAA, ICWS, ICSOC, AAMAS, and MICCAI. He regularly serves as PC for prestigious conferences (\emph{i.e.,} KDD, IJCAI, and CIKM) and reviewers for IEEE/ACM Transactions Journals. His research interests include big data analysis, data mining, and crowd science.
\end{IEEEbiography}
\vspace{-3em}

\newpage

\begin{center}
\textbf{\large Supplementary Materials}
\end{center}
{\appendix[Empirical Evaluations and Analysis] 
We reorganize the publicly available benchmark multivariate time series datasets, and select 9 commonly used datasets for evaluation. Then, we introduce the experimental setup and evaluation metrics used to assess the performance. Finally, we present a comprehensive comparison of the selected state-of-the-art models, with a particular focus on contrasting methods.

\subsection{Datasets}
\label{datasets}

In the field of multivariate time series analysis, there are mainly from 3 archives of datasets, including UCI\footnote{http://archive.ics.uci.edu/ml/datasets.php}, UEA\footnote{http://www.timeseriesclassification.com/dataset.php} and MTS\footnote{http://www.mustafabaydogan.com/multivariate-time-series-discretization-for-classification.html}. Particularly, it is important to note that there may be some overlap and discrepancies within these datasets. In this work, we carefully reorganize the multivariate time series datasets considering multiple aspects, including application domains, sequence lengths, number of samples, variables, and classes, to ensure our evaluation cover datasets with diverse characteristics. Our reorganization allows for a more systematic and comprehensive analysis of the datasets in relation to their specific characteristics.

\begin{itemize}
\item \textbf{UCI archive:} The UCI archive contains 85 multivariate time series datasets specifically curated for classification tasks. These datasets encompass a wide range of application fields, such as audio spectra classification, business, ECG/EEG classification, human activity recognition, gas detection, motion classification, etc. More details are listed in Table~\ref{tab:UCI}. 
\item \textbf{UEA archive:} The UEA archive collects 30 multivariate time series classification datasets that span various application domains, including audio spectra classification, ECG/EEG/MEG classification, human activity recognition, motion classification, etc. More details are listed in Table~\ref{tab:UEA}.
\item \textbf{MTS archive:} The MTS archive, also known as Baydogan's archive, consists of 13 multivariate time series datasets covering applications of audio spectra classification, ECG classification, human activity recognition, motion classification, etc. More details are listed in Table~\ref{tab:MTS}.
\end{itemize}

Additionally, we also prioritize datasets that have been widely adopted in existing research works. Considering the fact that MTS has lower usage and has significant overlap with UEA and UCI, we ultimately select 9 representative datasets primarily from the UCI and UEA archives for evaluation. These datasets include PhonemeSpectra (PS), DuckDuckGeese (DDG), EigenWorms (EW), PenDigits (PD), Epileptic Seizure Recognition (Epilepsy), FingerMovements (FM), Human Activity Recognition Using Smartphones (HAR), Smartphone and Smartwatch Activity and Biometrics Dataset (WISDM) and UniMiB SHAR (SHAR). They encompass various application fields, such as digits, images, and biology. For datasets from UEA, we utilize the pre-defined train-test split. For the remaining datasets from the UCI archive, we partition the data into 80\% for training and 20\% for testing. Detailed information about the selected datasets can be found in Table~\ref{tab:selected}.

\begin{table}[b]
    \centering
    \setlength{\tabcolsep}{4.5pt}
    \begin{tabular}{l|l|l|llll}
    \toprule[1.2pt]
   Task & Archive & Dataset & $N$ & $V$ & Classes & Length   \\
   \midrule
      {ASC} & UEA & PS &  6,668 & 11 & 39  & 217 \\ 
\cline{2-7}   &UEA &   DDG &  100 & 1,345 & 5 & 270  \\ 
  \hline
      {MC} & UEA & EW & 259 & 6 & 5 & 17,984 \\
\cline{2-7}   & UCI/UEA & PD & 10,992 & 2 & 10 & 8  \\
  \hline
      {EC} & UCI &  Epilepsy & 12,500 & 1 & 2 & 178  \\
\cline{2-7}   & UEA & FM  & 416 & 28 & 2 & 50  \\
      \hline
      {HAR} & UCI & HAR &  10,299 & 9 & 6 & 128  \\
\cline{2-7}   & UCI & WISDM & 4,091 & 3 & 6 & 200 \\
\cline{2-7}   & Others & SHAR & 11,771 & 3 & 17 & 151\\
   \bottomrule[1.2pt]
    \end{tabular}
    \caption{Statistical information of the selected time series classification datasets for evaluation. ASC refers to Audio Spectra Classification, MC refers to Motion Classification, EC refers to ECG/EEG/EMG/MEG  Classification, and HAR refers to Human Activity Recognition. $N$ denotes the number of samples and $V$ denotes the number of variables. Classes refer to the number of classes in each dataset and Length refers to the length of each time series.}
    \label{tab:selected}
\end{table}

\begin{table*}
\renewcommand\arraystretch{0.95}
\setlength{\tabcolsep}{9.7pt}
\scriptsize
  \centering
    \begin{tabular}{llllll}
      \toprule[1.2pt]
      Dataset & Application & $N$ & $V$ \\ 
      \midrule
      FMA: A Dataset For Music Analysis & Audio Spectra Classification & 106,574  & 518  \\
      Japanese Vowels & Audio Spectra Classification & 640  & 12   \\
      Spoken Arabic Digit & Audio Spectra Classification & 8,800  & 13   \\
      BAUM-1 & Audio Spectra Classification & 1,184    & - \\
      BAUM-2 & Audio Spectra Classification & 1,047    & - \\
      3W dataset & Business & 1984  & 8   \\
      CNNpred: CNN-based stock market prediction using a diverse set of variables & Business & 1,985  & 84   \\
      Dow Jones Index & Business & 750  & 16   \\
      ISTANBUL STOCK EXCHANGE & Business & 536  & 8   \\
      Machine Learning based ZZAlpha Ltd. Stock Recommendations 2012-2014 & Business & 314,080    &  - \\
      Online Retail & Business & 541,909  & 8   \\
      Online Retail II & Business & 1,067,371  & 8   \\
      MHEALTH Dataset & ECG Classification & 120  & 23   \\
      EEG Eye State & EEG Classification & 14,980  & 15   \\
      EEG Steady-State Visual Evoked Potential Signals & EEG Classification & 9,200  & 16   \\
      Epileptic Seizure Recognition & EEG Classification & 11,500  & 179   \\
      EMG data for gestures & EMG Classification & 30,000  & 6   \\
      EMG Physical Action Data Set & EMG Classification & 10,000  & 8   \\
      sEMG for Basic Hand movements & EMG Classification & 3,000  & 2,500   \\
      Robot Execution Failures & Failure Detection & 463  & 90   \\
      AI4I 2020 Predictive Maintenance Dataset & Failure Detection & 10,000  & 14   \\
      Breath Metabolomics & Gas Detection & 104  & 1,656   \\
      Gas Sensor Array Drift Dataset at Different Concentrations & Gas Detection & 13,910  & 129   \\
      Gas sensor array exposed to turbulent gas mixtures & Gas Detection & 180  & 150,000   \\
      Gas sensor array temperature modulation & Gas Detection & 4,095,000  & 20   \\
      Gas sensor array under dynamic gas mixtures & Gas Detection & 4,178,504  & 19   \\
      Gas sensor array under flow modulation & Gas Detection & 58  & 120,432   \\
      Gas sensor arrays in open sampling settings & Gas Detection & 18,000  & 1,950,000   \\
      Gas sensors for home activity monitoring & Gas Detection & 919,438  & 11   \\
      Ozone Level Detection & Gas Detection & 2,536  & 73   \\
      Twin gas sensor arrays & Gas Detection & 640  & 480,000   \\
      Activities of Daily Living (ADLs) Recognition Using Binary Sensors & Human Activity Recognition & 2,747    &  -\\
      Activity Recognition from Single Chest-Mounted Accelerometer & Human Activity Recognition   &  - &  -\\
      Activity Recognition system based on Multisensor data fusion (AReM) & Human Activity Recognition & 42,240  & 6   \\
      Bar Crawl: Detecting Heavy Drinking & Human Activity Recognition & 14,057,567  & 3   \\
      Basketball dataset & Human Activity Recognition & 10,000  & 7   \\
      Daily and Sports Activities & Human Activity Recognition & 9,120  & 5,625    \\
      Daphnet Freezing of Gait & Human Activity Recognition & 237  & 9   \\
      Dataset for ADL Recognition with Wrist-worn Accelerometer & Human Activity Recognition &  3  & 14  \\
      Heterogeneity Activity Recognition & Human Activity Recognition & 43,930,257  & 16   \\
      Human Activity Recognition from Continuous Ambient Sensor Data & Human Activity Recognition & 13,956,534  & 37   \\
      Human Activity Recognition Using Smartphones & Human Activity Recognition & 10,299  & 561   \\
      Intelligent Media Accelerometer and Gyroscope (IM-AccGyro) Dataset & Human Activity Recognition & 800  & 9   \\
      Localization Data for Person Activity & Human Activity Recognition & 164,860  & 8   \\
      MEx & Human Activity Recognition & 6,262  & 710   & \\
      OPPORTUNITY Activity Recognition & Human Activity Recognition & 2,551  & 242   \\
      PAMAP2 Physical Activity Monitoring & Human Activity Recognition & 3,850,505  & 52   \\
      REALDISP Activity Recognition Dataset & Human Activity Recognition & 1,419  & 120   \\
      selfBACK & Human Activity Recognition & 26,136  & 6   \\
      Simulated Falls and Daily Living Activities Data Set & Human Activity Recognition & 3,060  & 138   \\
      Smartphone Dataset for Human Activity Recognition in Ambient Assisted Living (AAL) & Human Activity Recognition & 5,744  & 561   \\
      Smartphone-Based Recognition of Human Activities and Postural Transitions & Human Activity Recognition & 10,929  & 561   \\
      User Identification From Walking Activity & Human Activity Recognition  & - & -  \\
      Vicon Physical Action Data Set & Human Activity Recognition & 3,000  & 27   \\
      WISDM Smartphone and Smartwatch Activity and Biometrics Dataset & Human Activity Recognition & 15,630,426  & 6   \\
      BLE RSSI dataset for Indoor localization & Indoor localization  & 23,570  & 5   \\
      BLE RSSI Dataset for Indoor localization and Navigation & Indoor localization  & 6,611  & 15   \\
      Geo-Magnetic field and WLAN dataset for indoor localisation from wristband and smartphone & Indoor localization  & 153,540  & 25   \\
      Hybrid Indoor Positioning Dataset from WiFi RSSI, Bluetooth and magnetometer & Indoor localization  & 1,540  & 65   \\
      UJIIndoorLoc-Mag & Indoor localization  & 40,000  & 13   \\
      Kitsune Network Attack Dataset & Malware/Attack Identification & 27,170,754  & 115  \\
      URL Reputation & Malware/Attack Identification & 2,396,130  & 3,231,961   \\
      Detect Malware Types & Malware/Attack Identification & 7,107  & 280   \\
      Dynamic Features of VirusShare Executables & Malware/Attack Identification & 107,888  & 482   \\
      BitcoinHeistRansomwareAddressDataset & Malware/Attack Identification & 2,916,697  & 10   \\
      Character Trajectories & Motion Classification & 2,858  & 3   \\
      Gesture Phase Segmentation & Motion Classification & 9,900  & 50   \\
      Indoor User Movement Prediction from RSS data & Motion Classification & 13,197  & 4   \\
      Pedestrian in Traffic Dataset & Motion Classification & 4,760  & 14   \\
      WESAD (Wearable Stress and Affect Detection) & Motion Classification & 63,000,000  & 12   \\
      Australian Sign Language signs & Motion Classification & 6,650  & 15   \\
      Australian Sign Language signs (High Quality) & Motion Classification & 2,565  & 22   \\
      PEMS-SF & Others & 440  & 138,672   \\
      Parking Birmingham & Others & 35,717  & 4   \\
      Occupancy Detection & Others & 20,560  & 7   \\
      Synthetic Control Chart Time Series & Others & 600  & 6  \\
      Crop mapping using fused optical-radar data set & Others & 325,834  & 175   \\
      Absenteeism at work & Others & 740  & 21   \\
      Educational Process Mining (EPM): A Learning Analytics Data Set & Others & 230,318  & 13   \\
      Productivity Prediction of Garment Employees & Others & 1,197  & 15   \\
      Buzz in social media & Others & 140,000  & 77   \\
      Condition monitoring of hydraulic systems & Others & 2,205  & 43,680   \\
      Open University Learning Analytics dataset & Others  & - & -   \\
      Data for Software Engineering Teamwork Assessment in Education Setting & Others & 74  & 102   \\
      Behavior of the urban traffic of the city of Sao Paulo in Brazil & Others & 135  & 18  \\
      \bottomrule[1.2pt]
      \end{tabular}
\caption{The UCI Multivariate Time Series Classification Archive - 85 Datasets. $N$ denotes the number of samples and $V$ denotes the number of variables. Classes refer to the number of classes in each dataset and Length refers to the length of each time series.}
  \label{tab:UCI}
\end{table*}

\begin{table*}
\small
\setlength{\tabcolsep}{17.7pt}
\renewcommand\arraystretch{1.2}
  \centering
    \begin{tabular}{llllll}
      \toprule[1.2pt]
      Dataset & Application & $N$ & $V$ & Classes & Length \\
      \midrule
      DuckDuckGeese & Audio Spectra Classification & 100 & 1,345 & 5  & 270 \\
      Heartbeat & Audio Spectra Classification & 409 & 61 & 2  & 405 \\
      InsectWingbeat & Audio Spectra Classification & 50,000 & 200 & 10 & 30 \\
      JapaneseVowels & Audio Spectra Classification & 640 & 12 & 9  & 29 \\
      PhonemeSpectra & Audio Spectra Classification & 6,668 & 11 & 39  & 217 \\
      SpokenArabicDigits & Audio Spectra Classification & 8,798 & 13 & 10  & 93 \\
      AtrialFibrillation & ECG Classification & 30 & 2 & 3  & 640 \\
      StandWalkJump & ECG Classification & 27 & 4 & 3  & 2,500 \\
      FaceDetection & EEG/MEG Classification & 9,414 & 144 & 2  & 62 \\
      FingerMovements & EEG/MEG Classification & 416 & 28 & 2  & 50 \\
      HandMovementDirection & EEG/MEG Classification & 234 & 10 & 4  & 400 \\
      MotorImagery & EEG/MEG Classification & 378 & 64 & 2  & 3,000 \\
      SelfRegulationSCP1 & EEG/MEG Classification & 561 & 6 & 2  & 896 \\
      SelfRegulationSCP2 & EEG/MEG Classification & 380 & 7 & 2  & 1,152 \\
      BasicMotions & Human Activity Recognition & 80 & 6 & 4  & 100 \\
      Cricket & Human Activity Recognition & 180 & 6 & 12  & 1,197 \\
      Epilepsy & Human Activity Recognition & 275 & 3 & 4 & 206 \\
      ERing & Human Activity Recognition & 300 & 4 & 6 & 65 \\
      Handwriting & Human Activity Recognition & 1,000 & 3 & 26 & 152 \\
      Libras & Human Activity Recognition & 360 & 2 & 15 & 45 \\
      NATOPS & Human Activity Recognition & 360 & 24 & 6 & 51 \\
      RacketSports & Human Activity Recognition & 303 & 6 & 4  & 30 \\
      UWaveGestureLibrary & Human Activity Recognition & 440 & 3 & 8  & 315 \\
      ArticularyWordRecognition & Motion Classification & 575 & 9 & 25  & 144 \\
      CharacterTrajectories & Motion Classification & 2,858 & 3 & 20  & 182 \\
      EigenWorms & Motion Classification & 259 & 6 & 5  & 17,984 \\
      PenDigits & Motion Classification & 10,992 & 2 & 10  & 8 \\
      EthanolConcentration & Others & 524 & 3 & 4  & 1,751 \\
      LSST & Others & 4,925 & 6 & 14  & 36 \\
      PEMS-SF & Others & 440 & 963 & 7  & 144 \\
      \bottomrule[1.2pt]
      \end{tabular}
  \caption{The UEA Multivariate Time Series Classification Archive - 30 Datasets. $N$ denotes the number of samples and $V$ denotes the number of variables. Classes refer to the number of classes in each dataset and Length refers to the length of each time series. }
  \label{tab:UEA}
\end{table*}

\begin{table*}
\small
\setlength{\tabcolsep}{19pt}
\renewcommand\arraystretch{1.2}
  \centering
  \begin{tabular}{llllll}
      \toprule[1.2pt]
      Dataset & Application & $N$ & $V$ &  Classes & Length \\
      \midrule
      ArabicDigits & Audio Spectra Classification & 8,800 & 13 & 10 & 4 - 93 \\
      JapaneseVowels & Audio Spectra Classification & 640 & 12 & 9 & 7 - 29 \\
      ECG & ECG Classification & 200 & 2 & 2 & 39 - 152 \\
      Libras & Human Activity Recognition & 360 & 2 & 15 & 45 \\
      UWave & Human Activity Recognition & 4,478 & 3 & 8 & 315 \\
      AUSLAN & Motion Classification & 2,565 & 22 & 95 & 45 - 136 \\
      CharacterTrajectories & Motion Classification & 2,858 & 3 & 20 & 109 - 205 \\
      CMUsubject16 & Motion Classification & 58 & 62 & 2 & 127 - 580 \\
      KickvsPunch & Motion Classification & 26 & 62 & 2 & 274 - 841 \\
      WalkvsRun & Motion Classification & 44 & 62 & 2 & 128 - 1,918 \\
      NetFlow & Others & 1,337 & 4 & 2 & 50 - 997 \\
      PEMS & Others & 440 & 963 & 7 & 144 \\
      Wafer & Others & 1,194 & 6 & 2 & 104 - 198 \\
      \bottomrule[1.2pt]
      \end{tabular}
     \caption{The MTS Multivariate Time Series Classification Archive - 13 Datasets. $N$ denotes the number of samples and $V$ denotes the number of variables. Classes refer to the number of classes in each dataset and Length refers to the length of each time series.}
  \label{tab:MTS}
\end{table*}}

\subsection{Experimental Setup}
\label{sec:metric}
We conduct linear evaluation to assess the performance of state-of-the-art unsupervised learning models, with a particular focus on various contrastive methods, on 9 representative datasets mentioned in Section~\ref{datasets}. All models are implemented using the PyTorch framework, and the experimental evaluations are conducted on an NVIDIA GeForce RTX 3090 GPU for efficient computation. To ensure a fair comparison, we follow the standard linear benchmarking evaluation scheme employed in ~\cite{cpc, simclr,tstcc}. This involves freezing the representations pre-trained by each unsupervised model and attaching a linear classifier on top of the frozen representations. The linear classifier is a logistic regression classifier, which consists of a fully-connected layer followed by the softmax activation function. We disable gradient computation on the inputs to the linear classifier. The classification performance is measured by using the following 3 commonly used metrics: ACC (Accuracy), MF1 (macro-averaged F1 score) and $\kappa$ (Cohen's Kappa coefficient). All the models are trained for 200 epochs. The mean and standard deviation of empirical results obtained from 5 repeated experiments are reported for each metric. 

\begin{itemize}

\item ACC: Accuracy is the ratio of the number of correctly classified samples to the total number of samples, \emph{i.e.,} ACC $= \frac{\text{TP}+\text{TN}}{\text{TP}+\text{TN}+\text{FP}+\text{FN}}$, where TP, TN, FP, FN are true positive, true negative, false positive and false negative. 

\item MF1: Macro-averaged F1 score is calculated as the arithmetic mean of individual classes' F1 score, \emph{i.e.,} MF1 $=\frac{2\times \text{PR}}{\text{P}+\text{R}}$, where Precision P and Recall R are calculated by $\text{P} =\frac{\text{TP}}{\text{TP}+\text{FP}}, \text{R} =\frac{\text{TP}}{\text{TP}+\text{FN}}$.

\item $\kappa$: Cohen's Kappa coefficient measures the inter-annotator agreement that expresses the level of agreement between predicted and true labels on classification, \emph{i.e.,} $\kappa = \frac{\text{ACC}-p_e}{1-p_e}$, where $p_e=\frac{[(\text{TP}+\text{FN}) *(\text{TP}+\text{FP})+(\text{FP}+\text{TN})*(\text{FN}+\text{TN})]}{ N^{2}}$ is the hypothetical probability of chance agreement, and $N$ denotes the total number of samples.

\end{itemize}

\begin{table*}[t]
\renewcommand\arraystretch{1.05}
\setlength{\tabcolsep}{2.6pt}
\begin{tabular}{l|l|ccc|ccc|ccc}
\toprule[1.2pt]  
\multirow{2}{*}{Category}     & \multirow{2}{*}{Model}      &     & HAR &  &     & WISDM &  &     & Epilepsy  \\
\cline{3-11}
&     & ACC & MF1   & $\kappa$ & ACC & MF1      & $\kappa$ & ACC & MF1     & $\kappa$  \\
\midrule
\multirow{2}{*}{Deep   Clustering}  & DeepCluster &  69.95$\pm$2.47 & 67.41$\pm$2.19 & 63.43$\pm$2.20      &     72.89$\pm$2.38 & 70.39$\pm$3.16 & 68.07$\pm$3.45   & 86.07$\pm$1.84 & 80.99$\pm$1.97 & 76.43$\pm$2.18  \\
& IDFD   &   67.07$\pm$1.62	&  64.12$\pm$1.45 & 60.87$\pm$1.40	& 68.14$\pm$1.82 & 	64.78$\pm$1.91 & 	61.08$\pm$1.93	& 79.56$\pm$1.73	&  74.84$\pm$1.52	& 70.24$\pm$1.56\\\midrule
\multirow{2}{*}{Reconstruction-based}      & TimeNet     &   74.61$\pm$2.42 & 73.64$\pm$2.51 & 70.35$\pm$2.73 & 79.23$\pm$2.76 & 71.82$\pm$2.73 & 70.49$\pm$3.04 & 89.32$\pm$1.39 & 85.71$\pm$1.43 & 75.76$\pm$1.92  \\
& Deconv      &  71.53$\pm$1.65 & 70.76$\pm$1.84 & 62.72$\pm$2.39 & 75.76$\pm$2.30 & 69.45$\pm$2.79 & 66.28$\pm$3.21 & 85.44$\pm$1.92 & 79.85$\pm$2.08 & 76.01$\pm$2.57   \\\midrule
Self-supervised  & TimeGAN   &  73.57$\pm$1.96 & 70.04$\pm$1.94 & 66.71$\pm$2.30 & 79.22$\pm$2.61 & 72.43$\pm$2.88 & 69.26$\pm$3.03 & 89.79$\pm$1.24 & 86.55$\pm$1.36 & 74.83$\pm$1.77 \\
~~-Adversarial   & TimeVAE      &    70.69$\pm$2.07  & 65.61$\pm$2.41 & 61.45$\pm$2.89 & 74.46$\pm$2.97 & 67.93$\pm$3.41 & 64.10$\pm$3.94 & 86.91$\pm$1.17 & 82.40$\pm$1.41 & 76.98$\pm$2.56\\
\cline{2-11}
~~-Predictive   & EEG-SSL     &   65.34$\pm$1.63 & 	 63.75$\pm$1.37 & 	57.20$\pm$1.22 
&   70.67$\pm$1.11	&  66.50$\pm$0.93	&  62.44$\pm$1.07	&  93.72$\pm$0.45 	&  89.15$\pm$0.93	& 
77.65$\pm$1.64 \\
& TST    &   70.73$\pm$2.29 & 66.36$\pm$2.03 &  63.14$\pm$2.74 &    76.68$\pm$2.85 & 70.06$\pm$2.94 & 67.30$\pm$3.41   &  91.21$\pm$0.88 & 87.64$\pm$1.33 & 87.49$\pm$1.75     \\
\cline{2-11}
~~-Contrastive  & SimCLR      &  81.06$\pm$2.35 & 80.62$\pm$2.31 & 77.25$\pm$2.82   &   83.04$\pm$4.21 &75.83$\pm$6.47 & 75.15$\pm$6.27    &  93.00$\pm$0.57 & 88.09$\pm$0.97 & 76.27$\pm$1.93  \\
& BYOL   &  89.46$\pm$0.17 & 89.31$\pm$0.17 & 87.33$\pm$0.20   &   87.84$\pm$0.38 & 84.02$\pm$0.77 & 82.43$\pm$0.57    &     98.08$\pm$0.09 & 96.99$\pm$0.15 & 93.99$\pm$0.30  \\
& SwAV   &   68.81$\pm$1.50 &  66.69$\pm$1.56 &  62.41$\pm$1.81    &     73.44$\pm$1.28 &  53.90$\pm$3.56 &  59.75$\pm$2.29     &   94.30$\pm$0.85 &  90.80$\pm$1.37 &  81.62$\pm$2.73    \\
& PCL    &  74.49$\pm$1.95 & 65.02$\pm$1.69 & 71.96$\pm$2.12   &   69.47$\pm$1.51 & 61.75$\pm$3.29 & 56.36$\pm$2.57    &     89.93$\pm$1.34 & 87.68$\pm$1.42 & 85.78$\pm$1.88  \\
& TS-TCC      &  89.22$\pm$0.70 & 89.23$\pm$0.76 & 87.03$\pm$0.85   &   81.48$\pm$0.98 & 69.17$\pm$2.25 & 73.13$\pm$1.33    &      97.19$\pm$0.18 &  95.47$\pm$0.31 & 90.94$\pm$0.61     \\
& T-Loss      &   91.06$\pm$0.94 &  90.94$\pm$0.96 &  89.26$\pm$1.13  &  91.48$\pm$1.05 &  88.79$\pm$1.53 &  87.79$\pm$1.51     &    96.94$\pm$0.20 &  95.20$\pm$0.30 &  90.41$\pm$0.61     \\
\bottomrule[1.2pt]   
\end{tabular}
\caption{Comparisons between different categories of unsupervised learning methods. ACC refers to accuracy, MF1 refers to macro-averaged F1 score, and $\kappa$ refers to Cohen's kappa coefficient. }
\label{tab:allcategory}
\end{table*}

\subsection{Results and Discussions}
We choose 3 time series datasets for conducting comparisons among various categories of unsupervised learning methods. Following this analysis, due to the promising results and growing popularity of contrastive learning methods over other unsupervised learning methods, we allocate special emphasis on contrastive learning methods covering multiple levels of contrast in the following subsections.

\subsubsection{A comparative analysis of different categories of unsupervised learning methods}
The comparisons between different categories of unsupervised learning methods on 3 selected datasets are presented in Table~\ref{tab:allcategory}. In general, self-supervised learning approaches outperform deep clustering and reconstruction-based methods. Such performance gain indicates that self-supervised models offer improved representation learning capabilities. This can be attributed to the broader range of pretext tasks, such as predicting missing elements, temporal order, or context, which encourage the model to learn rich and meaningful representations. This diversity of tasks enables the model to capture different aspects of the data, leading to more comprehensive and robust representations. Particularly, contrastive methods have demonstrated promising results in most datasets as they distinguish samples from multiple views without the need for explicit labels. Such discrimination allows the learned representations to be more effective in capturing meaningful differences between data samples. Instead, adversarial methods such as TimeGAN~\cite{timegan} and TimeVAE~\cite{timevae} assume specific data distributions, require iterative optimization and often are more challenging to converge. These methods can be computationally expensive and require larger amounts of data for training. Whereas, contrastive methods are less sensitive to the data distribution compared to adversarial methods and are generally easier to train. Additionally, predictive methods including EEG-SSL~\cite{eegssl} and TST~\cite{tst}, tend to perform worse than most contrastive methods. This is because predictive methods rely on a large number of given contexts to accurately predict part of the data, which can be challenging in practice. Conversely, contrastive methods can make better use of the available data by focusing on differentiating between a smaller number of similar and dissimilar data pairs. 
\label{652}

\begin{table}[h]
\renewcommand\arraystretch{0.94} 
\setlength{\tabcolsep}{3pt}
\centering
\begin{tabular}{l|l|l|c|c|c}
\toprule[1.2pt]
Dataset    & Level   & Model  & ACC     & MF1     & $\kappa$    \\
\midrule
\multirow{9}{*}{Epilepsy}  & \multirow{3}{*}{Instance} & SimCLR & 93.00$\pm$0.57   & 88.09$\pm$0.97   & 76.27$\pm$1.93     \\
&   & BYOL   & 98.08$\pm$0.09   & 96.99$\pm$0.15   & 93.99$\pm$0.30  \\
&   & CPC    & 96.61$\pm$0.43   & 94.44$\pm$0.69   & 88.67$\pm$1.22   \\
\cline{2-6}
& \multirow{3}{*}{Prototype} & SwAV   & 94.30$\pm$0.85   & 90.80$\pm$1.37   & 81.62$\pm$2.73    \\
&   & PCL    & 89.93$\pm$1.34   & 87.68$\pm$1.42   & 85.78$\pm$1.88   \\
&   & MHCCL  & 97.85$\pm$0.49   & 95.44$\pm$0.82   & 91.08$\pm$1.57       \\
\cline{2-6}
& \multirow{3}{*}{Temporal} & TS2Vec & 96.32$\pm$0.23   & 94.27$\pm$0.37   & 88.54$\pm$0.74     \\
&   & TS-TCC & 97.19$\pm$0.18   & 95.47$\pm$0.31   & 90.94$\pm$0.61  \\
&   & T-Loss & 96.94$\pm$0.20   & 95.20$\pm$0.30   & 90.41$\pm$0.61  \\
\midrule
\multirow{9}{*}{HAR}    & \multirow{3}{*}{Instance}   & SimCLR & 81.06$\pm$2.35   & 80.62$\pm$2.31   & 77.25$\pm$2.82    \\
& & BYOL   & 89.46$\pm$0.17   & 89.31$\pm$0.17   & 87.33$\pm$0.20   \\
& & CPC    & 83.85$\pm$1.51   & 83.27$\pm$1.66   & 79.76$\pm$1.90      \\
\cline{2-6}
& \multirow{3}{*}{Prototype}& SwAV   & 68.81$\pm$1.50   & 66.69$\pm$1.56   & 62.41$\pm$1.81    \\
&  & PCL    & 74.49$\pm$1.95   & 65.02$\pm$1.69   & 71.96$\pm$2.12   \\
& & MHCCL  & 91.60$\pm$1.06   & 91.77$\pm$1.11   & 89.90$\pm$1.27   \\
\cline{2-6}
& \multirow{3}{*}{Temporal} & TS2Vec & 90.47$\pm$0.66   & 90.46$\pm$0.64   & 89.15$\pm$0.79     \\
& & TS-TCC & 89.22$\pm$0.70   & 89.23$\pm$0.76   & 87.03$\pm$0.85   \\
& & T-Loss & 91.06$\pm$0.94   & 90.94$\pm$0.96   & 89.26$\pm$1.13   \\
\midrule
\multirow{9}{*}{SHAR}   & \multirow{3}{*}{Instance}  & SimCLR & 67.22$\pm$1.76   & 53.05$\pm$2.42   & 63.84$\pm$1.97     \\
&  & BYOL   & 67.00$\pm$0.98   & 59.30$\pm$1.30   & 63.62$\pm$1.10   \\
& & CPC    & 68.57$\pm$1.40   & 62.44$\pm$1.77   & 61.62$\pm$0.96   \\
\cline{2-6}
& \multirow{3}{*}{Prototype} & SwAV   & 57.49$\pm$2.75   & 58.82$\pm$2.53   & 58.89$\pm$3.50    \\
&  & PCL    & 56.28$\pm$1.47   & 49.42$\pm$1.62   & 51.78$\pm$1.65  \\
& & MHCCL  & 83.42$\pm$1.76   & 78.45$\pm$2.09   & 80.58$\pm$1.84   \\
\cline{2-6}
& \multirow{3}{*}{Temporal}  & TS2Vec & 82.94$\pm$2.91   & 77.89$\pm$2.95   & 78.94$\pm$3.20   \\
& & TS-TCC & 70.80$\pm$0.75   & 64.28$\pm$0.72   & 67.74$\pm$0.85   \\
& & T-Loss & 80.88$\pm$3.94   & 77.06$\pm$3.44   & 79.65$\pm$4.35    \\
\midrule
\multirow{9}{*}{WISDM}    & \multirow{3}{*}{Instance} & SimCLR & 83.04$\pm$4.21   & 75.83$\pm$6.47   & 75.15$\pm$6.27    \\
& & BYOL   & 87.84$\pm$0.38   & 84.02$\pm$0.77   & 82.43$\pm$0.57    \\
&  & CPC    & 80.35$\pm$0.98   & 73.24$\pm$1.21   & 72.23$\pm$1.07   \\
\cline{2-6}
& \multirow{3}{*}{Prototype}  & SwAV   & 73.44$\pm$1.28   & 53.90$\pm$3.56   & 59.75$\pm$2.29  \\
&  & PCL    & 69.47$\pm$1.51   & 61.75$\pm$3.29   & 56.36$\pm$2.57   \\
&  & MHCCL  & 93.60$\pm$1.06   & 91.70$\pm$1.10   & 90.96$\pm$1.08   \\
\cline{2-6}
& \multirow{3}{*}{Temporal} & TS2Vec & 92.33$\pm$1.05   & 90.27$\pm$1.07   & 90.36$\pm$1.49    \\
&  & TS-TCC & 81.48$\pm$0.98   & 69.17$\pm$2.25   & 73.13$\pm$1.33    \\
& & T-Loss & 91.48$\pm$1.05   & 88.79$\pm$1.53   & 87.79$\pm$1.51   \\
\bottomrule[1.2pt]
\end{tabular}
\caption{A comparative analysis of different levels of contrast on 4 datasets. ACC refers to accuracy, MF1 refers to macro-averaged F1 score, and $\kappa$ refers to Cohen's kappa coefficient.}
\label{tab:contrastive_other}
\end{table}

\begin{table}[h]
\setlength{\tabcolsep}{3pt}
\renewcommand\arraystretch{0.94} 
\centering
\begin{tabular}{l|l|l|c|c|c}
\toprule[1.2pt]
Dataset & Level& Model & ACC & MF1 & $\kappa$  \\
\midrule
\multirow{9}{*}& \multirow{3}{*}{Instance}  & SimCLR & 40.00$\pm$0.60 & 39.01$\pm$0.69 & 25.00$\pm$0.74 \\
 && BYOL & 51.60$\pm$1.20 & 52.04$\pm$1.34 & 39.50$\pm$1.50   \\
 && CPC & 44.72$\pm$2.19 & 42.90$\pm$3.45 & 30.77$\pm$4.16  \\
 \cline{2-6}
 & \multirow{3}{*}{Prototype}& SwAV & 52.40$\pm$3.44 & 50.95$\pm$4.46 & 40.50$\pm$4.30  \\
{DDG} && PCL & 48.62$\pm$3.56 & 46.86$\pm$3.97 & 35.67$\pm$5.74  \\
 && MHCCL & 50.76$\pm$3.19 & 49.52$\pm$3.54 & 39.50$\pm$4.27  \\
 \cline{2-6}
  & \multirow{3}{*}{Temporal} & TS2Vec & 52.00$\pm$3.27 & 51.02$\pm$3.64 & 40.00$\pm$4.08\\
 && TS-TCC & 38.67$\pm$5.96 & 35.51$\pm$6.48 & 23.33$\pm$7.45  \\
 && T-Loss & 54.50$\pm$4.33 & 52.70$\pm$5.48 & 43.13$\pm$5.41  \\
\midrule
\multirow{9}{*} & \multirow{3}{*}{Instance}& SimCLR & 49.20$\pm$0.98 & 43.08$\pm$4.90 & -0.69$\pm$2.29  \\
& & BYOL & 49.60$\pm$1.11 & 49.38$\pm$1.12 & -0.49$\pm$2.22  \\
 && CPC & 49.73$\pm$1.45 & 44.76$\pm$1.69 & -0.27$\pm$2.70  \\
 \cline{2-6}
 & \multirow{3}{*}{Prototype}& SwAV & 44.22$\pm$2.19 & 44.19$\pm$2.20 & 8.47$\pm$4.40  \\
{FM} & &PCL & 50.44$\pm$1.95 & 43.68$\pm$1.30 & 1.19$\pm$3.89  \\ 
 & &MHCCL & 52.09$\pm$1.45 & 50.51$\pm$2.06 & 17.87$\pm$3.39  \\
  \cline{2-6}
 & \multirow{3}{*}{Temporal}& TS2Vec & 50.00$\pm$1.63 & 49.99$\pm$1.64 & -0.01$\pm$3.30  \\
 & &TS-TCC & 50.17$\pm$1.60 & 49.07$\pm$2.10 & 0.33$\pm$4.58  \\
 & &T-Loss & 50.50$\pm$1.72 & 50.33$\pm$1.76 & 4.01$\pm$6.65  \\
\midrule
\multirow{9}{*} & \multirow{3}{*}{Instance}& SimCLR & 93.35$\pm$0.17 & 93.27$\pm$0.17 & 92.61$\pm$0.19  \\
 && BYOL & 94.93$\pm$0.08 & 94.96$\pm$0.07 & 94.37$\pm$0.09   \\
 && CPC & 92.03$\pm$0.32 & 91.00$\pm$0.35 & 90.04$\pm$0.36  \\
  \cline{2-6}
 & \multirow{3}{*}{Prototype}& SwAV & 78.02$\pm$2.86 & 76.86$\pm$3.06 & 71.10$\pm$3.19  \\
 {PD} && PCL & 86.18$\pm$1.25 & 83.36$\pm$1.11 & 80.69$\pm$1.01  \\ 
 && MHCCL & 98.69$\pm$0.41 & 98.71$\pm$0.55 & 97.43$\pm$0.72  \\
  \cline{2-6}
 & \multirow{3}{*}{Temporal}& TS2Vec & 97.83$\pm$0.24 & 97.80$\pm$0.25 & 97.59$\pm$0.27  \\
 && TS-TCC & 97.44$\pm$0.23 & 97.45$\pm$0.23 & 97.16$\pm$0.26  \\
 && T-Loss & 97.86$\pm$0.52 & 97.87$\pm$0.52 & 97.63$\pm$0.57  \\
\midrule
\multirow{9}{*} & \multirow{3}{*}{Instance}& SimCLR & 15.14$\pm$0.30 & 12.64$\pm$0.33 & 12.91$\pm$0.31  \\
 && BYOL & 12.08$\pm$0.21 & 11.62$\pm$0.19 & 9.77$\pm$0.21   \\
 && CPC & 14.31$\pm$0.78 & 14.32$\pm$0.77 & 9.95$\pm$0.84   \\
  \cline{2-6}
& \multirow{3}{*}{Prototype} & SwAV & 11.66$\pm$0.64 & 8.92$\pm$0.58 & 9.33$\pm$0.65  \\
 {PS} && PCL & 8.92$\pm$1.04 & 8.96$\pm$1.24 & 6.53$\pm$1.49   \\ 
 && MHCCL & 15.43$\pm$0.79 & 14.29$\pm$0.76 & 12.63$\pm$0.79   \\
  \cline{2-6}
  & \multirow{3}{*}{Temporal}& TS2Vec & 15.69$\pm$0.90 & 15.69$\pm$0.95 & 13.47$\pm$0.92 \\
 && TS-TCC & 10.90$\pm$0.76 & 9.16$\pm$0.65 & 8.56$\pm$0.78   \\
 && T-Loss & 18.60$\pm$0.25 & 18.15$\pm$0.28 & 16.46$\pm$0.25  \\
\midrule
\multirow{9}{*}  & \multirow{3}{*}{Instance}& SimCLR & 60.87$\pm$1.84 & 55.62$\pm$4.69 & 47.99$\pm$3.27 \\
 && BYOL & 79.39$\pm$0.84 & 75.12$\pm$1.05 & 71.84$\pm$1.19   \\
 && CPC & 67.03$\pm$1.38 & 63.06$\pm$1.38 & 58.26$\pm$1.53  \\
  \cline{2-6}
 & \multirow{3}{*}{Prototype} & SwAV & 43.75$\pm$1.80 & 35.16$\pm$2.31 & 21.82$\pm$2.27 \\
 {EW} & &PCL & 43.28$\pm$1.06 & 40.02$\pm$2.01 & 22.58$\pm$1.59   \\
 & &MHCCL & 79.10$\pm$2.21 & 76.01$\pm$1.45 & 69.53$\pm$1.90  \\
  \cline{2-6}
  & \multirow{3}{*}{Temporal} & TS2Vec & 82.80$\pm$2.52 & 81.59$\pm$3.73 & 80.74$\pm$3.41\\
 && TS-TCC & 73.21$\pm$1.44 & 64.33$\pm$0.81 & 63.41$\pm$1.17   \\
 && T-Loss & 75.00$\pm$3.39 & 67.85$\pm$3.64 & 65.12$\pm$4.91  \\
 \bottomrule[1.2pt]
\end{tabular}
\caption{A comparative analysis of different levels of contrast on UEA datasets. ACC refers to accuracy, MF1 refers to macro-averaged F1 score, and $\kappa$ refers to Cohen's kappa coefficient.}
\label{tab:contrastive_uea}
\end{table}

\subsubsection{A comparative analysis of performing different levels of contrast}
We conducted a specific comparison and analysis of contrastive methods focusing on different levels of contrast. Table~\ref{tab:contrastive_uea} shows the comparative analysis at different levels of contrast on UEA datasets, and the results on 4 other datasets are shown in Table~\ref{tab:contrastive_other}. From the results, it can be observed that contrastive learning models such as TS2Vec~\cite{ts2vec}, TS-TCC~\cite{tstcc} and T-Loss~\cite{tloss}, which emphasize the impact at the temporal level, achieve better results compared to methods that focus on instance-level or prototype-level contrast. However, prototype-level contrastive learning models such as SwAV~\cite{swav} and PCL~\cite{pcl}, which have shown great success in image data, do not perform as well in time series data. These methods usually employ flat clustering algorithms, which are capable of capturing only a single hierarchy of semantic clusters. Nonetheless, the result obtained from one clustering iteration cannot be consistently relied upon as there is no assurance of their accuracy, rendering them inadequate. In this case, it is worth noting that MHCCL~\cite{mhccl} demonstrates superior performance as it incorporates hierarchical clustering into the construction of contrastive pairs, which helps address the limitation of flat clustering algorithms.

\subsubsection{A comparative analysis of using different backbones}

Table~\ref{tab:backbone} presents a comparative analysis of employing different backbones in the SimCLR model for unsupervised time series representation learning. Typically, we find that most models summarized in Table~\ref{tab:summary_instance}, Table~\ref{tab:summary_prototype} and Table~\ref{tab:summary_temporal} utilize ResNets~\cite{resnet} as the backbone and achieve satisfactory performance. Therefore, we compare the results of using ResNet-18 and ResNet-50 as backbones. Furthermore, we evaluate InceptionTime~\cite{inceptiontime}, which is specifically designed for time series data by incorporating the temporal dependencies. As shown in Table~\ref{tab:backbone}, InceptionTime outperforms the general-purpose ResNets in most datasets. This observation highlights the capability of InceptionTime to identify both local and global shape patterns (\emph{i.e.,} low-level and high-level features) in time series data. InceptionTime proves to be exceptionally fitting for time series data due to its utilization of 1D convolutions with multiple kernel sizes in parallel. This approach facilitates efficient feature extraction on time series data across various scales, making it well-suited to capture diverse granularities of semantics in time series data. In contrast, ResNets primarily rely on 2D convolutions and are optimized for image data, which may not yield the same level of effectiveness when applied to time series data analysis.

\begin{table}[t]
\centering
\renewcommand\arraystretch{0.94} 
\begin{tabular}{l|c|c|c|c}
\toprule[1.1pt]
Dataset      &   Metric  & ResNet-18   & ResNet-50   & InceptionTime \\
\midrule     
\multirow{3}{*}{HAR}     & ACC    & 78.70$\pm$2.27  & 81.06$\pm$2.35 & 82.25$\pm$1.94    \\
& MF1    & 78.44$\pm$2.19  & 80.62$\pm$2.31 & 80.56$\pm$2.28    \\
& $\kappa$ & 76.21$\pm$2.43  & 77.25$\pm$2.82  & 78.39$\pm$2.65    \\
\midrule     
\multirow{3}{*}{WISDM}   & ACC    & 86.94$\pm$1.92  & 83.04$\pm$4.21 & 87.46$\pm$1.87    \\
& MF1    & 81.65$\pm$3.26  & 75.83$\pm$6.47  & 82.14$\pm$2.85    \\
& $\kappa$ & 80.91$\pm$2.95  &  75.15$\pm$6.27  & 81.62$\pm$2.24    \\
\midrule   
\multirow{3}{*}{Epilepsy}   & ACC    & 93.42$\pm$0.26  & 93.00$\pm$0.57  & 93.68$\pm$0.41    \\
& MF1    & 88.72$\pm$0.54  &88.09$\pm$0.97  & 88.80$\pm$0.64    \\
& $\kappa$ & 77.54$\pm$1.06  & 76.27$\pm$1.93   & 77.89$\pm$1.39    \\
\bottomrule[1.1pt]    
\end{tabular}
\caption{The classification performance of using different backbones in SimCLR. ACC refers to accuracy, MF1 refers to macro-averaged F1 score, and $\kappa$ refers to cohen's kappa coefficient.}
\label{tab:backbone}
\end{table}

\subsubsection{A comparative analysis of employing different augmentation selection strategies}
Table~\ref{tab:augmentation} presents the classification accuracy of employing different data augmentation selection strategies. We evaluate 4 strategies in 3 popular contrastive learning models on 3 datasets. As shown in Table~\ref{tab:augmentation}, we compare the original strategy (Original) used in each model with 3 alternative augmentation selection strategies. The specified one (Specified) follows the implementation in TS-TCC~\cite{tstcc} which employs simple yet efficient transformations that can fit any time series data to create a strong and a weak view. Therefore, the specified strategy is equivalent to the original one in TS-TCC model. Both the aforementioned original and specified augmentation selection strategies are developed based on domain expert knowledge and empirical results. Additionally, we also compare with 2 automatic selection strategies including RandAugment~\cite{randaugment} and W-Augment~\cite{waugment}. Nevertheless, it is noteworthy that the advancements achieved through the utilization of automatic augmentation selection strategies have shown minimal improvement. In some cases, the performance does not surpass that of the original performance. Consequently, developing automatic augmentation selection strategies specifically tailored for time series data or optimizing the combinations of various transformations remains an open problem for boosting contrastive learning.
\begin{table}[h]
\centering
\renewcommand\arraystretch{0.94} 
\begin{tabular}{l|c|c|c|c}
\toprule[1.1pt]
Dataset    &   Strategy  & SimCLR   & TS-TCC  & BYOL \\
\midrule     
\multirow{4}{*}{HAR}     & Original     & 81.83$\pm$2.18 & 89.22$\pm$0.70 & 89.36$\pm$2.09  \\
& Specified    & 81.06$\pm$2.35 & 89.22$\pm$0.70 &  89.46$\pm$0.17   \\
& RandAugment  & 81.02$\pm$2.37 & 87.94$\pm$2.14 & 89.90$\pm$2.37    \\
& W-Augment  & 81.27$\pm$1.66 & 87.57$\pm$1.69 &  89.35$\pm$1.96   \\
\midrule     
\multirow{4}{*}{WISDM}   & Original     & 83.71$\pm$3.48 & 81.48$\pm$0.98 & 87.91$\pm$2.42 \\
& Specified    & 83.04$\pm$4.21 & 81.48$\pm$0.98 &  87.84$\pm$0.38    \\
& RandAugment  & 82.78$\pm$2.69  &  82.32$\pm$2.36 &  88.03$\pm$2.99   \\
& W-Augment  & 84.30$\pm$1.85 & 82.47$\pm$1.87 &  87.84$\pm$2.25   \\
\midrule   
\multirow{4}{*}{Epilepsy}   & Original   & 92.97$\pm$1.62 & 97.19$\pm$0.18 & 97.64$\pm$1.23 \\
& Specified    & 93.00$\pm$0.57 &  97.19$\pm$0.18 & 98.08$\pm$0.09    \\
& RandAugment  & 93.26$\pm$2.17 & 98.04$\pm$0.64 &  98.06$\pm$0.91   \\
& W-Augment  & 92.19$\pm$1.79 & 97.52$\pm$0.92 &  97.67$\pm$0.79   \\
\bottomrule[1.1pt]    
\end{tabular}
 \caption{The classification accuracy of employing different data augmentation selection strategies.}
 \label{tab:augmentation}
\end{table}

\vfill

\end{document}